\documentclass{bmvc2k}

\usepackage{times}
\usepackage{epsfig}
\usepackage{graphicx}
\usepackage{amsmath}
\usepackage{amssymb}
\usepackage{multirow}
\usepackage{tabularx}
\usepackage{afterpage}
\usepackage{capt-of}

\usepackage{comment}
\usepackage{colortbl}
\usepackage{color}

\usepackage{chapterbib}

\definecolor{Konrad}{RGB}{255, 0, 0}
\definecolor{Audrey}{RGB}{255, 0, 255}
\definecolor{Christoph}{RGB}{0, 0, 255}
\definecolor{Maros}{RGB}{0, 255, 0}

\makeatletter

\def\eg{\emph{e.g}\bmvaOneDot} 
\def\ie{\emph{i.e}\bmvaOneDot} \def\Ie{\emph{I.e}\bmvaOneDot}
\def\cf{\emph{cf}\bmvaOneDot} 
\def\etc{\emph{etc}\bmvaOneDot} 
\def\wrt{w.r.t\bmvaOneDot}

\newcommand{\norm}[2][]{\|{#2}\|_{{#1}}}
\newcommand{\scal}[2]{\langle #1,#2 \rangle}

\newcommand{\si}[2]{\textrm{si}_{{#1}}^{{#2}}}

\newcommand{\Eq}{Eq.\@\xspace}
\newcommand{\Eqs}{Eqs.\@\xspace}
\newcommand{\Fig}{Fig.\@\xspace}

\newcommand{\Sec}{Sec.\@\xspace}
\DeclareMathOperator*{\argmin}{arg\,min}

\DeclareMathOperator*{\sign}{sign}

\newcommand{\parspc}{\vspace{1mm}}

\definecolor{myGreen}{rgb}{0.2,0.6,0.}
\newcommand{\myparagraph}[1]{\vspace{0.5em}\noindent\textbf{#1}}

\newcommand{\todo}[1]{\textcolor{blue}{\footnotesize\sf \textbf{ToDo:} #1}\xspace}

\newcommand{\invisible}[1]{}
\newcommand{\RNum}[1]{{\bf (\lowercase\expandafter{\romannumeral #1\relax})}}
\newcommand{\trans}{^{\!\mathsf{T}}}

\renewcommand{\div}{\nabla\!\cdot\!}
\newcommand{\dx}{\mathrm{d}x}
\newcommand{\dz}{\mathrm{d}z}

\title{Semantic 3D Reconstruction with Finite Element Bases}

\addauthor{Audrey Richard}{audrey.richard@geod.baug.ethz.ch}{1,$\dagger$}
\addauthor{Christoph Vogel}{christoph.vogel@icg.tugraz.at}{2,$\dagger$}
\addauthor{Maro\v{s} Bl\'aha}{maros.blaha@geod.baug.ethz.ch}{1}
\addauthor{Thomas Pock}{thomas.pock@icg.tugraz.at}{2,3}
\addauthor{Konrad Schindler}{konrad.schindler@geod.baug.ethz.ch}{1}

\addinstitution{
 Photogrammetry \& Remote Sensing\\
 ETH Zurich, Switzerland
}
\addinstitution{
 Institute of Computer Graphics \& Vision\\
 TU Graz, Austria
}
\addinstitution{
 Austrian Institute of Technology
}

\addauthorship{
\vspace{1.5cm}
$^{\dagger}$ shared first authorship
}

\runninghead{Richard, Vogel, et al.}{Semantic 3D Reconstruction with FEM Bases}

\begin{document}

\graphicspath{{./justPaperArxiv/}}

\maketitle

\begin{abstract}
We propose a novel framework for the discretisation of multi-label
problems on arbitrary, continuous domains.
Our work bridges the gap between general FEM discretisations, and
labeling problems that arise in a variety of computer vision tasks,
including for instance those derived from the generalised Potts
model.
%
%
Starting from the popular formulation of labeling as a convex
relaxation by functional lifting, we show that FEM discretisation
is valid for the most general case, where the regulariser is
anisotropic and non-metric.
While our findings are generic and applicable to different vision
problems, we demonstrate their practical implementation in the context
of semantic 3D reconstruction, where such regularisers have proved
particularly beneficial.
The proposed FEM approach leads to a smaller memory footprint as well
as faster computation, and it constitutes a very simple way to enable
variable, adaptive resolution within the same model.

\end{abstract}

\vspace{-0.7em}
\section{Introduction}
\label{sec:intro}
\vspace{-0.3em}
%
A number of computer vision tasks, such as segmentation, multiview
reconstruction, stitching and inpainting, can be formulated
as \emph{multi-label problems on continuous domains}, by functional
lifting \cite{PockCBC10,Cremers11,Lellmann11,ChambolleCP12,NieuwenhuisTC13}.
%
A recent example is semantic 3D reconstruction
(\eg \cite{HaneZCAP13,BlahaVRWPS16}), which solves the following
problem: Given a set of images of a scene, reconstruct both its 3D
shape and a segmentation into semantic object classes.
%
The task is particularly challenging, because the evidence is
irregularly distributed in the 3D domain; but it also possesses a
rich, anisotropic prior structure that can be exploited.
Jointly reasoning about shape and class allows one to take into
account class-specific shape priors (\eg, building walls should be
smooth and vertical, and vice versa smooth, vertical surfaces are
likely to be building walls), leading to improved reconstruction
results.
So far, models for the mentioned multi-label problems, and in
particular for semantic 3D reconstruction, have been limited to
axis-aligned discretisations. Unless the scenes are aligned with the
coordinate axes, this leads to an unnecessarily large number of
elements. Moreover, since the evidence is (inevitably) distributed
unevenly in 3D, it also causes biased reconstructions.
Thus, it is desirable to adapt the discretisation to the scene content
(as often done for purely geometric surface reconstruction,
\emph{e.g.}~\cite{Labatut07}).

Our formulation makes it possible to employ a finer tesselation in
regions that are likely to contain a surface, exploiting the fact that
both high spatial resolution and high numerical precision are only
required in those regions.
Our discretisation scheme leads to a smaller memory footprint and faster
computation, and it constitues a very simple technique to allow for
arbitrary adaptive resolution levels within the same problem.
\Ie, we can refine or coarsen the discretisation as appropriate, to adapt
to the scene to be reconstructed.
While our scheme is applicable to a whole family of finite element
bases, we investigate two particularly interesting cases: Lagrange
(P1) and Raviart-Thomas elements of first order.
%
%
We further show that the grid-based voxel discretisation is a special
case of our P1 basis, such that minimum energy solutions of
``identical'' discretisations (same vertex set) are equivalent.
%
%

\begin{figure}[t]
\begin{center}
   \vspace{-1mm}
   \includegraphics[width=1\linewidth]{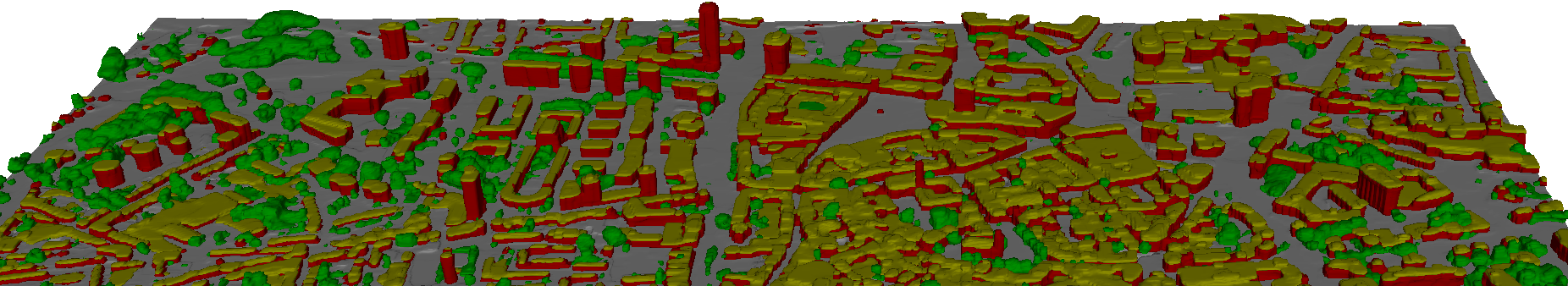}
\end{center}
   \vspace{-6.25mm} \caption{Semantic 3D model, estimated from aerial views with our FEM
   method.}
\label{fig:3d_reconstruction}
\vspace{-6mm}
\end{figure}

\vspace{-0.7em}
\section{Related Work}
\label{sec:related}
\vspace{-0.3em}

Since the seminal work \cite{Curless1996} volumetric reconstruction
from image data has evolved remarkably
\cite{Kazhdan:2006:PSR,VogiatzisETC07,Furu:2010:PMVS,LiuC10,Cremers-Kolev-pami11,kbc-pami11,KostrikovHL14,Ulusoy2016}.
Most methods use depth maps or 2.5D range scans for evidence
\cite{Zach2007_Range,zach2008fast}, represent the scene via an
indicator or signed distance function in the volumetric domain, and
extract the surface as its zero level set, \eg,
\cite{Lorensen1987,treece1999mtet}.
%

Joint estimation of geometry and semantic labels, which had earlier
been attempted only for single depth maps \cite{Ladicky10}, has
recently emerged as a powerful extension of volumetric 3D
reconstruction from multiple views
\cite{HaneZCAP13,Bao13,Kundu14,Savinov15,vineet2015icra,BlahaVRWPS16,Ulusoy2017CVPR}.
A common trait of these works is the integration of depth estimates
and appearance-based labeling information from multiple images, with
class-specific regularisation via shape priors.
%
%

Multi-label problems are in general NP-hard, but under certain
conditions on the pairwise interactions, the original non-convex
problem can be converted into a convex one via functional lifting and
subsequent relaxation, \eg \cite{ChambolleCP12}.
This construction was further extended to anisotropic
(direction-dependent) regularisers \cite{Strekalovskiy11}.
Moreover, \cite{ZachHP14} also relaxed the requirement that the
regulariser forms a metric on the label set, yet its construction can
only be applied after discretisation \cite{Lellmann11}.
In this paper, we consider the relaxation in its most general form
\cite{ZachHP14}, but are not restricted to it.
%
%
The latter construction is also the basis to the model of
\cite{HaneZCAP13}, whose energy model we adapt for our semantic 3D
reconstruction method.
Their voxel-based formulation can be seen as a special case of our
discretisation scheme.

%
For (non-semantic) surface reconstruction, several authors prefer a
data-dependent discretisation, normally a Delaunay tetrahedralisation
of a 3D point cloud \cite{Labatut07,Jancosek11,vu12}.
The occupancy states of the tetrahedra are found by discrete (binary)
labeling, and the final surface is composed of the triangles that
separate different labels.
%
%
Loosely speaking, our proposed methodology can be seen either as an
extension of \cite{HaneZCAP13} to arbitrary simplex partitions of the
domain; or as an extension of \cite{Labatut07} to semantic
(multi-label) reconstruction.


We note that regular voxel partitioning of the volume leads to
a high memory consumption and computation time.
Yet, we are essentially reconstructing a 2D manifold in 3D space,
and this can be exploited to reduce run-time and memory footprint.
\cite{Kundu14} use an octree instead of equally sized voxels to adapt
to the local density of the input data.
\cite{BlahaVRWPS16} go one step further and propose an adaptive
octree, where the discretisation is refined on-the-fly, during
optimisation.
In our framework the energy is independent of the discretisation, it
can thus be combined directly with such an adaptive procedure.
%

Also volumetric fusion via signed distance
functions \cite{Newcombe2011} benefits from irregular tesselations of
3D space, \eg, octrees \cite{Steinbruecker-etal-icra14} or
hashmaps \cite{Niessner2013}.
In contrast to our work, these target real-time reconstruction and
refrain from global optimisation, instead locally fusing depth maps.
Their input normally is a densely sampled, overcomplete RGB-D
video-stream, whereas we deal with noisy and incomplete inputs.
To achieve high-quality reconstructions in our setting, we
incorporate semantic information, leading to a multi-label problem.
%
%
%

Our work is based on the finite element method (FEM), \eg
\cite{reddy2005introduction,Brezzi1991}. Introduced by Ritz
\cite{Ritz1909} more than a century ago, and refined by Galerkin and
Courant \cite{courant1943}, FEM serves to numerically solve
variational problems, by partitioning the domain into finite,
parametrised elements.
%
%
%
%
In computer vision FEM has been applied in the context of level-set
methods \cite{Weber2004} and for Total Variation \cite{Bartels12}.
To our knowledge, we are the first to apply it to multi-labeling.

\vspace{-0.7em}
\section{Method}
\vspace{-0.3em}
\label{sec:method}
The multi-labeling
problem \cite{ChambolleCP12,Strekalovskiy11,Lellmann11,ZachHP14} in
the domain $\Omega\subset \mathbb{R}^d$ is defined by finding $m$
labeling functions $x^i:\Omega \rightarrow \{0,1\}, i=1\ldots m$ as
the solution of:
\vspace{-0.1cm}
\begin{equation}\label{eq:multiLabelProblem}
\inf_{x^i} \sum_{i=1}^m \int_\Omega \rho^i(z) x^i(z) \dz + J(x^i),
\quad\textrm{s.\ t. } \sum_{i=1}^{m} x^i(z)=1 \;\forall z\in\Omega,
\vspace{-0.1cm}
\end{equation}
where $\rho$ models the data term for a specific label at location
$z\in\Omega$ and $J$ denotes a convex regularisation functional that enforces
the spatial consistency of the labels.
One prominent example is to chose $J:= \norm{\cdot}{_2}$, known as Total Variation,
which penalises the perimeter of the individual regions
\cite{ChambolleCP12,NieuwenhuisTC13}.
Note that in the two-label case (Potts model), this relaxation is exact
after thresholding with any threshold from the open unit
interval \cite{ChambolleCP12}.
%
Although we are ultimately interested in non-metric regularisation, we
start with the continuous, anisotropic model \cite{Strekalovskiy11},
and postpone the extension to the non-metric case to Sec.~\ref{subsec:nonmetric}.

\vspace{-0.9em}
\subsection{Convex Relaxation}
\vspace{-0.3em}
The continuous model allows for an anisotropic regulariser in $J$:
label transitions can be penalised on the area of the shared surface,
as well as on the surface normal direction.
This is achieved with problem-specific 1-homogeneous functions that
emerge from convex sets, so called Wulff-shapes.
%
A relaxation of $x^i(z)\in\{ 0,1\}$ to $x^i(z)\in[0,1]$ then leads to a convex
energy, which can be written as the following saddle point problem, with primal functions $x$ and dual functions $\lambda$:
\vspace{-0.1cm}
\begin{equation}\label{eq:saddle}
\min_{x^i} \max_{\lambda^i} \sum_i\!\!\! \int_\Omega\!\!\!\! \rho^i\!(z) x^i(z) \!\!+\!\! \langle x^i(z), \div \lambda^i(z) \rangle \dz,
\;\;
\textrm{s.t. } \lambda^i(z)\!-\!\lambda^j(z) \in W^{ij}\!\!
, 
\sum_{i=1}^{m} \!x^i(z)\!\!=\!\!1 , 
x^i(z)\!\geq\!0 .
\vspace{-0.1cm}
\end{equation}
%
The constraints have to be fulfilled for all $z\in\Omega$.
In addition to the primal variables $x^i$, we have introduced the dual
vector-field $\lambda^i:\mathbb{R}^d\rightarrow \mathbb{R}^d$, whose pairwise
differences are constrained to lie in the convex sets (Wulff-shapes) $W^{ij}$.
By letting these shapes take an anisotropic form, one can then encode
scene structure, \eg \cite{Strekalovskiy11,HaneZCAP13}.  For our
problem we demand Neumann conditions at the boundary of $\Omega$, \ie
$\langle \lambda^i,\nu \rangle =0,\forall z\in\partial\Omega$,
because the scene will continue beyond our domain ($\nu$ is the normal
of the domain boundary $\partial \Omega$).
%

\vspace{-0.7em}
\subsection{Finite Element Spaces}
\vspace{-0.3em}
Here, we can only informally introduce the basic idea of FEM and
explain its suitability for problems of the form
\eqref{eq:saddle}. We refer to textbooks
\cite{Larson2013FEM,Brezzi1991,Duran2008} for a deeper and
formal treatment.
%

One way to solve ~\eqref{eq:saddle} is to approximate it at
a finite number of regular grid points, using finite
differences.
%
FEM instead searches for a solution in a finite-dimensional vector space;
this \emph{trial} space is a subspace of the space in which the exact
solution is defined.
To that end, one chooses a suitable basis for the \emph{trial} space,
with basis functions of finite support, as well as an
appropriate \emph{test} function space.
%
FEM methods then find approximate solutions to variational problems by
identifying the element from the \emph{trial} space that is orthogonal to all
functions of the \emph{test} function space.
For our saddle-point problem, we can instead identify the \emph{trial} space with
our primal function space and the \emph{test} space with its dual counterpart.
%
Now, we can apply the same principles, and after discretisation our
solution corresponds to the continuous solution defined
by the respective basis.
%
%
%
%
%
As \eqref{eq:saddle} is already a relaxation of the original
problem \eqref{eq:multiLabelProblem}, we do not present an analysis of
convergence at this point.
Instead, the reader is referred to \cite{Bartels12} for an
introduction to this somewhat involved topic.
%

In order to choose a space with good approximation properties and
suitable basis functions, we tesselate our domain into simplices.
%
More formally, we define $M\!=\!\{F,V,S\}$ to be a simplex mesh with
vertices $v\!\in\! V, v\!\in\!\mathbb{R}^d$, faces $f\!\in\! F$ defined by $d$,
and simplices $s\!\in\! S$, defined by $d\!+\!1$ vertices that
partition $\Omega$: $\cup_k s_k=\Omega, s_l\cap s_k\!=\!f_{l,k}\!\in\! F$
-- \ie two adjacent simplices share only a single face.
%
In this work, for a specific set of vertices $V$, we select M to be
the corresponding Delaunay tetrahedralisation of $\Omega$
and only consider explicit bases. In particular, we
focus on the Lagrangian (P1) basis, which we use in the following to
derive our framework; and on the Raviart-Thomas (RT) basis.
Details for the latter are given in the supplementary material.
%
The main difference between them is that P1 leads to piecewise linear
solutions, which must be thresholded, while RT leads to a constant
labeling function per simplex, similar to discrete MAP solutions on CRFs.
We note that constant labeling can lead to artefacts, such that the
adaptiveness of the FEM model becomes even more important.

The idea of both derivations is similar:
\emph{(i)} select a basis for our primal (P1) or dual (RT) variable set,
\emph{(ii)} find a suitable form via the divergence theorem and Fenchel duality,
\emph{(iii)} extend to the non-metric case, following a principle we term ''label mass preservation''.
%
\vspace{-0.7em}
\subsection{Lagrange Elements}
\begin{figure}[tb]
\begin{center}
   \includegraphics[width=0.85\linewidth]{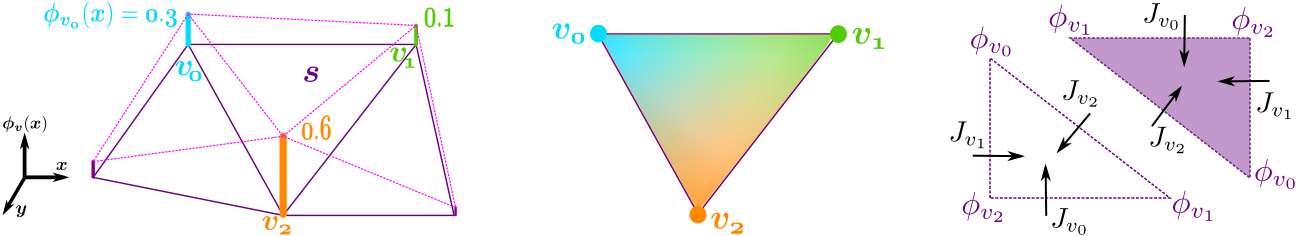}
\end{center}
   \vspace{-5mm} \caption{\textit{Left:} Illustration of P1 basis function shape. \textit{Middle:} Scalar field defined as a convex combination of basis coefficients. \textit{Right:} Gradient definition in a simplex (\ref{eq:gradient}).}
\label{fig:lagrange_elements}
\vspace{-4mm}
\end{figure}

\vspace{-0.3em}
%
%
The Lagrange $\textrm{P}^k(M)$ basis functions describe a
\emph{conforming} polynomial basis of
order $k\!+\!1$ on our simplex mesh $M$, \ie its elements
belong to the Hilbert space of differentiable function with
finite Lebesgue measure on the domain $\Omega$:
$\textrm{P}^k(M)\subset H^1(\Omega):= \{ p\in L^2(\Omega),\nabla p\in(L^2(\Omega)^d) \}$.
%
%
We are interested in the Lagrange basis of first order, $\textrm{P}^1(M)$:
\vspace{-.125cm}
\begin{align}
P^1(M) \!\!:=\! \{\! p\!:\Omega\!\rightarrow\! \mathbb{R} | p \!\in\! C(\Omega), \,p(x)\!:=\!\sum_{s\in S} c_s\trans x + d_s, c_s\!\in\!\mathbb{R}^d, d_s\!\in \!\mathbb{R},
\textrm{ if } x\!\in\! s \textrm{ and } 0 \textrm{ else}\} .
\vspace{-.125cm}
\end{align}
We construct our linear basis with functions that are defined
for each vertex $v$ of a simplex $s$ and can be described in
local form with barycentric coordinates:
\vspace{-0.175cm} 
\begin{equation}\label{eq:bariz}
p^{1}_{s,v}(x)   := \alpha_v\; \textrm{with } x=\sum_{v\in s} \alpha_v v, \; \sum_{v\in s} \alpha_v=1, \; \alpha_v\geq0 \quad \textrm{if}\; x\in s \;\textrm{and}\; 0 \;\textrm{else}.
\vspace{-0.125cm}
\end{equation}
In each simplex, one can define a scalar field $\phi_s(x)\in\mathbb{R}$ and
compute a gradient in this basis that will be constant
per simplex $s$ (\textit{cf.} Fig. \ref{fig:lagrange_elements}):
\goodbreak\noindent
\vspace{-0.125cm}
\begin{equation}\label{eq:gradient}
\phi_s(x) := \sum_{v\in s} \phi_v p^{1}_{s,v} \; \textrm{ and } \nabla \phi_s = \sum_{v\in s} \phi_v J_{v},
\vspace{-0.125cm}
\end{equation}
with coefficients $\phi_v\in \mathbb{R}$. $J_{v}\in\mathbb{R}^d$
denotes a vector that is normal to the face $f_v$ opposite node $v$,
has length $\frac{|f_v|}{|s|d}$, and points towards the simplex centre.
$|f_v|$ denotes the area of the face $f_v$,
and $|s|$
the volume of the simplex $s$
(\cf \Fig\ref{fig:lagrange_elements} and supplementary material).
%

%

\vspace{-0.7em}
\subsection{Discretisation}
\label{subsec:discretisation}
\vspace{-0.3em}
To apply our Lagrange basis to \eqref{eq:saddle} we first
make use of the divergence theorem:
\vspace{-0.075cm}
\begin{equation}\label{eq:p1_div}
\int_\Omega \langle x^i(z), \div \lambda^i(z) \rangle \dz =
\;
\int_\Omega \langle \nabla x^i(z), \lambda^i(z) \rangle \dz -
\underbrace{\int_{\partial\Omega} \langle \nu(z), \lambda^i(z) \rangle \dz}_{=0} .
\vspace{-0.25cm}
\end{equation}
The latter summand vanishes by our choice of $\lambda$.
Our approach for a discretisation in the Lagrange basis is to
choose the labeling function $x^i\in P^1(M)$.
This implies that our dual space consists of constant vector-fields
per simplex: $\lambda^i_{s}\in\mathbb{R}^d$.
To fulfill the constraint set in \eqref{eq:saddle} we have to verify
that, per simplex, the $\lambda^i_{s}$ lie in the respective Wulff-shape.
%
The simplex constraints on the $x^i$ have to be modeled per vertex.
According to \eqref{eq:bariz}, the labeling functions are convex
combinations of their values at the vertices and thus stay within the
simplex.
%

We also have to convert the continuous data costs $\rho^i$ into a
cost per vertex $\rho^i_v$, which can be achieved by
convolving the continuous cost with the respective basis function:
$\rho^i_v := \int_\Omega \sum_{s\in\mathcal{N}(v)} \phi_{s}(x) \rho^i(x)\dx$.
In practice, the integral can be computed by sampling $\rho$.
%
%
Integrating the right hand side in \eqref{eq:p1_div} over the simplex
$s$ leads to a weighting with its volume $|s|$ and the
energy \eqref{eq:saddle} in the discrete setting becomes:
\vspace{-0.1cm}
\begin{equation}\label{eq:pd1_discrete}
\min_{x^i}\max_{\lambda^i}
\sum_{v,i} \rho^i_v x_v^i \!+\! \sum_{s,i} |s| \langle \nabla x^i, \lambda_s^i \rangle \;\; \textrm{s.t. }
(\!\lambda_s^i-\lambda_s^j\!)\! \in\! W^{ij} \;\forall i\!<\!j,\; s\in S
, 
\sum_{i=1}^{m} \!x_v^i\!\!=\!\!1, \; x_v^i\!\geq\!0 \;\forall v\!\in\! V. 
\vspace{-0.1cm}
\end{equation}
%
%
%
%

\subsection{Non-metric extension}
\label{subsec:nonmetric}
\begin{figure}[t]
\begin{center}
   \includegraphics[width=0.9\linewidth]{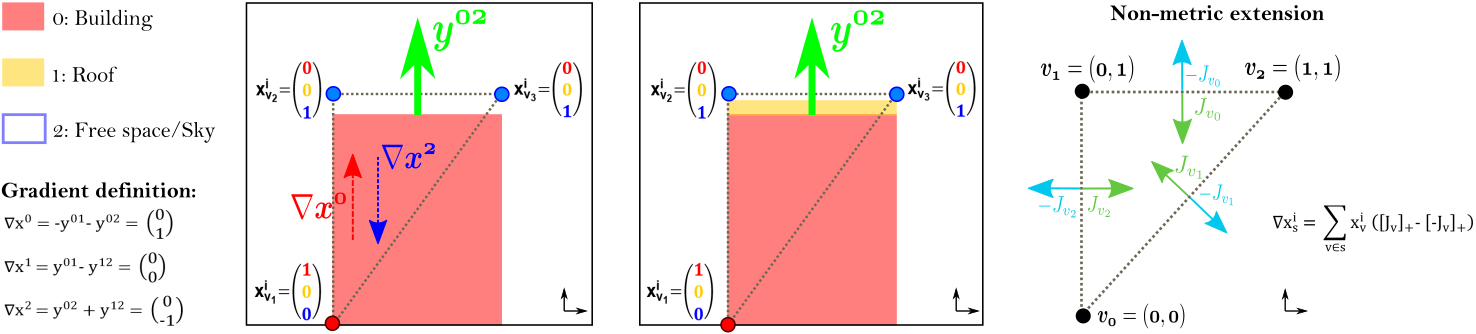}
\end{center}
   \vspace{-5mm} \caption{\textit{Left:} Without
	our non-metric extension,
	optimisation \wrt \eqref{eq:pd1_discrete} can lower transition
	costs by inserting another label (here 1 between 0 and
	2).  \textit{Right:} A solution is to split the
	gradients of the indicator functions and use direction-dependent
	variables $x^{ij}$. 
	}
%
\label{fig:non-metricness}
\vspace{-3mm}
\end{figure}

To start with, we note that a non-metric model does not exist in the
continuous case \cite{maggi2012sets} and our extension works
only \emph{after} the discretisation into the FEM basis.  Please refer
to the supplementary material for an in-depth discussion.
Note that our label set of semantic classes does not have a natural
order (in contrast to, \eg, stereo depth or denoised brightness); and
also the direction-dependent regulariser is unordered and does not
induce a metric cost.
%
%
%
%
%
%
To allow for non-metric regularisation we transform the constraint set
($\lambda^i-\lambda^j\in W^{ij}$), by introducing auxiliary
variables $z^{ij}$ and Lagrange multipliers $y^{ij}$, and use
Fenchel-Duality:
%
%
%
\vspace{-0.1cm}
\begin{equation}\label{eq:tranform1}
\max_{\lambda_s^{i},z_s^{ij}}\min_{y_s^{ij}}
\sum_{i<j}
\langle (\lambda_s^i-\lambda_s^j)-z^{ij}, y_s^{ij} \rangle - \delta_{W^{ij}}(z_s^{ij})
=
\max_{\lambda_s^{i}}\min_{y_s^{ij}}
\sum_{i<j}
- \langle (\lambda_s^i-\lambda_s^j), y_s^{ij} \rangle + || y_s^{ij} ||_{W^{ij}}.
\vspace{-0.1cm}
\end{equation}
The dual functions of the indicator functions for the convex
sets $W^{ij}$ are 1-homogeneous, of the form
$|| \cdot ||_{W^{ij}} \!:=\! \sup_{w\in W^{ij}} w\trans \cdot$.
%
%
%
Recall that our label costs are not metric:
$\forall i\!<\!j\!<\!k: |y^{ij}|_{W^{ij}} + |y^{jk}|_{W^{jk}} \geq |y^{ik}|_{W^{ik}}$,
does not hold.
It was shown \cite{ZachHP14} that a regulariser of the form
\eqref{eq:tranform1} transforms any non-metric cost to the metric case.
Figure \ref{fig:non-metricness} shows an example.  Here, an
expensive transition between labels 0 and 2 will be replaced by
two cheaper transitions 0--1 and 1--2.
To prevent this, we replace the $y^{ij}$ with direction dependent variables
$x^{ij}$: 
%
We rearrange \eqref{eq:tranform1}
and combine the first summand 
with the regulariser
from \eqref{eq:pd1_discrete} to arrive at the following equations
(for now ignoring $|s|$):
\vspace{-0.1cm}
\begin{equation*}
\sum_s \sum_i\langle \nabla x^i, \lambda_s^i \rangle
-  \sum_i \langle \lambda_s^i, \sum_{j\neq i} \left(y_s^{ij}[i<j]-y_s^{ji}[i>j] \right)\rangle, 
\vspace{-0.1cm}
\end{equation*}
with $[\cdot]$ denoting Iverson brackets.
Let $x^{ij}:= [y^{ij}]_+$ and $x^{ji}:= [-y^{ji}]_+$,
where $[\cdot]_+:=\max(0,\cdot)$. Expanding the gradient
\eqref{eq:gradient} we get, per  simplex $s$,
\vspace{-0.1cm}
\begin{equation}
\sum_i \sum_{v\in s} \lambda^i_s x_v^i( [J_{v}]_+ - [-J_{v}]_+)
- \langle \lambda_s^i, \sum_{j:i\neq j} (x^{ij}-x^{ji}) \rangle,
\vspace{-0.14cm} 
\end{equation}
which we analyse further to achieve non-metric costs.
%
It was observed in \cite{ZachHP14} that the $x^{ij}\in\mathbb{R}^d$
can be interpreted as encoding the ``label mass'' that transitions
from label $i$ to label $j$ in a specific direction.
%
Positivity constraints (by definition) on the $x^{ij}$ avoid
the transport of negative label mass.
%
%
%
To anchor transport of mass on the actual mass of label $i$ present at
a vertex, we introduce the variables $x^{ii}$ for the mass that
remains at label $i$, and split the above constraints into two separate sets
with the help of additional dual variables $\theta$:
\vspace{-0.075cm}
\begin{equation}\label{eq:nonMetric}
\lambda^i_s (\sum_{v\in s} x_v^i [ J_{v}]_+ - \sum_j x_s^{ij}) +
\theta^i_s  (\sum_{v\in s} x_v^i [-J_{v}]_+ - \sum_j x_s^{ji}) +
\sum_{i,j} \delta_{\geq0}(x_s^{ij}).
\vspace{-0.075cm}
\end{equation}

\noindent
Note that this construction is only possible because our
elements (simplices) are of strictly positive volume,
in contrast to zero sets in $\Omega$ \wrt the Lebesgue measure.
%
%
Finally, we can write down our discrete energy in the Lagrange basis
defined on the simplex mesh $M$:
\vspace{-0.001cm}
\begin{equation}\label{eq:final}
\begin{split}
\min_{x^i, x^{ij}} \max_{\lambda^i,\theta^i}
&
\sum_{v\in V} \sum_i \rho^i_v x_v^i + \delta_{\Delta}(x_v^i) + \sum_{i<j} \sum_{s\in S} |s| || x_s^{ij}-x_s^{ji} ||_{W^{ij}} +
\\
&
\sum_s \sum_i \theta^i_s  (\sum_{v\in s} x_v^i [-J_{v}]_+ - \sum_j x_s^{ji})
+
\sum_s \sum_i \lambda^i_s (\sum_{v\in s} x_v^i [ J_{v}]_+ - \sum_j x_s^{ij}) +
\sum_{i,j} \delta_{\geq0}(x_s^{ij}), \raisetag{3.5\baselineskip}
\end{split}
\vspace{-0.075cm}
\end{equation}
where we have moved the weighting with $|s|$ from the constraint set
to the regulariser, and denote by $\delta_{\Delta}(\cdot)$ the
indicator function of the unit simplex.

\vspace{-0.075cm}
\section{Semantic Reconstruction Model}\label{subsec:implementation}
\begin{figure}[tb]
\begin{center}
   \vspace{-1mm}
   \includegraphics[width=0.95\linewidth]{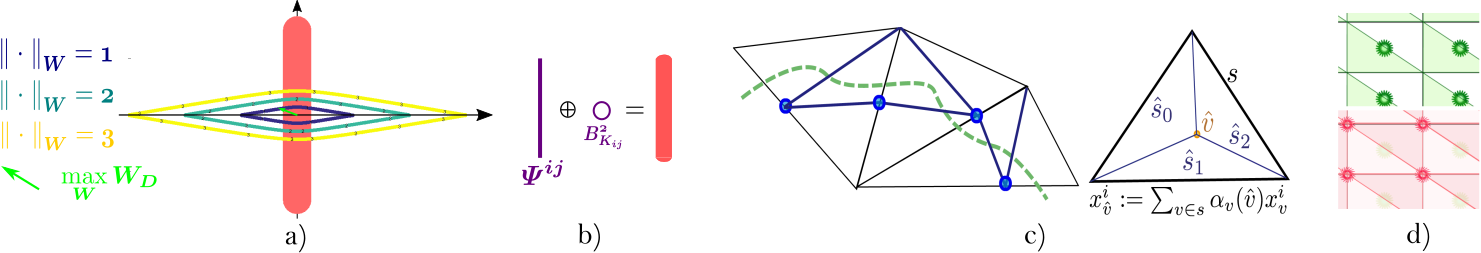}
\end{center}
   \vspace{-5mm}
\caption[Optional caption for list of figures]{
 \emph{(a):} Wulff-shape (\emph{red}) with isolines.
 \emph{(b):} Minkowski sum of two Wulff-shapes.
 \emph{(c):} Simplices are split after inserting new vertices (blue) close to the surface (green).
	\emph{Right:} Initialisation of vertices after refinement. 
\emph{(d):} Finite differences on a regular grid
  (\cite{HaneZCAP13}) only cover constraints in \emph{green} areas,
	the P1 basis covers all of the domain $\Omega$.
}
\label{fig:wulff_shape}
\vspace{-3.5mm}
\end{figure}

A prime application scenario for our FEM multi-label energy model
\eqref{eq:final} is 3D semantic reconstruction.
%
%
%
In particular, we focus on an urban scenario and let our labeling
functions encode \emph{freespace} ($i=1$), \emph{building wall},
\emph{roof}, \emph{vegetation} or \emph{ground}.
Objects that are not explicitly modeled are collected in an
extra \emph{clutter} class.
We define the data cost $\rho$ at a 3D-point $x\in\Omega$ as in
\cite{HaneZCAP13}: project
$x$ into the camera views $c\in\mathcal{C}$, and retrieve the
corresponding depth $\hat{d}_c(x)$ and class likelihoods
$\sigma_c^i(x)$ from the image space.
The $\sigma^i$ are obtained from a MultiBoost classifier. For the depth we
look at the difference between the actual distance $d_c(x)$ to the
camera and the observed depth: $d(x,c):=d_c(x)-\hat{d}_c(x)$.
For the \emph{freespace} label we always set the cost to 0, for $i\neq
1$ we define:
%
\vspace{-0.075cm}
\begin{equation}\label{eq:datacost}
\rho^i(x)\!:=\!\sum_{c\in\mathcal{C}}\! \sigma_c^i(x)[(k\!-1\!)\epsilon\leq d(x,c) \leq k\epsilon] +
\beta [|d(x,c)|\leq k\epsilon] \sign(d(x,c)).
\vspace{-0.075cm}
\end{equation}
%
%
This model assumes independence of the per-pixel observations, and
exponentially distributed inlier noise in the depthmaps, bounded by a
parameter $k\epsilon$ (k=3 in practice). It is essentially a
continuous version of \cite{HaneZCAP13}, see that paper for details.
%
%
%
The parameter $\epsilon$ sets a lower bound for the minimal height
of the simplices in the tesselation, and thus defines the target
resolution.
%
The discretisation of the data cost involves a convolution with the
respective basis functions, which can be approximated via sampling.
Please refer to the supplementary material for details.
%
%
%
%
The Wulff-shapes $W^{ij}$ in \eqref{eq:final} are given as the
Minkowski sum of the $L_2$-Ball,
$B^2_{\kappa^{ij}}:=\{x\in\mathbb{R}^d| \|x\|_2 \leq \kappa^{ij}\}$
and an anisotropic shape $\Psi^{ij}$:
$W^{ij}:= \Psi^{ij} \oplus B^2_{\kappa^{ij}}$.
In the isotropic part, $\kappa^{ij}$ contains the neighbourhood statistics
of the classes. The anisotropic part $\Psi^{ij}$ models the likelihood
of a transition between classes $i$ and $j$ in a certain direction.
\Fig\ref{fig:wulff_shape} (a,b) shows an example.
For our case we prefer flat, horizontal surfaces at the
following label transitions: \emph{ground-freespace}, \emph{ground-building},
\emph{building-roof}, \emph{ground-vegetation} and \emph{roof-freespace}.
A second prior prefers vertical boundaries for the transitions
\emph{building-freespace} and \emph{building-vegetation}.  More details on
the exact form can be found in \cite{HaneZCAP13}.

The energy \eqref{eq:final} is already in primal-dual form, such that
we can apply the minimisation scheme of \cite{chambolle11}, with
pre-conditioning \cite{Pock11}.
That numerical scheme requires us to project onto shapes that are
Minkowski sums of convex sets.  In our case, the sets are simple and
the projection onto each shape can be performed in closed form.
We employ a Dykstra-like projection scheme \cite{Dykstra}, which
avoids storing additional variables and proves remarkably efficient,
see supplementary material.
We also project the labeling functions $x^i$ directly onto the unit
simplex \cite{WangC13a}.
In order to extract the transition surface, we employ a variant of marching
tetrahedra (triangles) \cite{treece1999mtet}, using the isolevel at $0.5$
for each label.
We conclude with two interesting remarks.
First, note that a tesselation with a regular grid \cite{HaneZCAP13} can be
seen as a simplified version of our discretisation in the $P^1$ Lagrange basis.
In \Fig \ref{fig:wulff_shape}\textcolor{red}{d} 
we consider the 2D case of the regular grid
used in \cite{HaneZCAP13}. Here, variables are defined at voxel level.
In its dual graph, the vertex set consists of the corners of the primal
grid cubes, leading to shifted indicator variables.  
Per vertex the data term is mainly influenced from the cost in its Voronoi area.
%
%
Similarly, \cite{HaneZCAP13} evaluates the data cost at grid centers,
approximately corresponding to integration within the
respective Voronoi-area of grid cell. 
Furthermore, taking finite differences in this regular case
corresponds to verifying constraints for 
only one of the two triangles (\Fig\ref{fig:wulff_shape}\textcolor{red}{d}). 
The supplementary material includes a more formal analysis.
%

Second, our formulation is adaptive, in the sense
of \cite{BlahaVRWPS16}: Hierarchical refinement of the tesselation can
only decrease our energy. Hence, our scheme is applicable when refining
the model on-the-fly.
We again must defer a formal proof to the supplementary material and
give an intuitive, visual explanation in \Fig\ref{fig:wulff_shape}\textcolor{red}{c }. 
%
Assume that  $(x^*, \lambda^*, \theta^*)$ is an optimal solution
for a certain triplet $M=\{F,V,S\}$.  Then a refined tesselation
$\hat{M}=\{\hat{F},\hat{V},\hat{S}\}$ can be found by introducing
additional vertices, \ie $V\subset\hat{V}$ (ideally on the label
transition surfaces).
To define a new set of simplices, we demand that no
faces are flipped, $\forall \hat{s}\in \hat{S}, \exists s\in S$ with
$\hat{s} \cap s = \hat{s}$. Then one can find a new variable set and
data cost $\hat{\rho}$ with the same energy: 
%
We initialise the new variables from the
continuous solution at the respective location, and find new
$\rho_{\hat{v}}$ by integration.
Subsequent minimisation in the refined mesh can only decrease the
energy.
The argument works in both ways: Vertices that have the same solution
as their adjacent neighbors can be removed without changing the
energy.
For now we stick to this simple scheme, future work might explore more
sophisticated ideas, \eg along the lines of \cite{Grinspun02}.

\vspace{-0.7em}
\section{Evaluation}
\vspace{-0.3em}
\label{sec:evaluation}
Before we present results on challenging real 3D data we
evaluate our method in 2D on a synthetic dataset.
All results are obtained with a multi-core implementation, on a 12-core,
3.5 GHz machine with 64GB RAM.
For clarity, we only present the Lagrange discretisation. We refer to
the supplementary material for an evaluation of the Raviart-Thomas
discretisation.
%

\textbf{Input Data.}
We create a synthetic 2D scene composed of 4 labels: \emph{free space,
building, ground} and \emph{roof}, surrounded by 17 virtual cameras.
To replace depth maps and class-likelihood images, we extract 2D
points on the
boundary ``surface'' and assign ground truth label costs to each point.
For the evaluation in 3D, we use three real-world aerial mapping data
sets.
Our method requires two types of input data: depthmaps and pixel-wise
class probabilities (\cf Sec.~\ref{subsec:implementation}).
Moreover, we build a \emph{control mesh} $M$ around the initially
predicted surface, to facilitate our FEM discretisation.  Ideally, the
\textit{control mesh} enwraps the true surface, using a finer meshing close
to it.
We densely evaluate the data cost at the vertices of a regular
\emph{data cost grid} and let each \emph{control vertex} accumulate the cost of
its nearest neighbours in that grid, to approximate an integration
over its Voronoi cell.
\parspc

%
%

\begin{figure}[tb]
\begin{minipage}{.575\textwidth}
  \centering
\includegraphics[width=70mm]{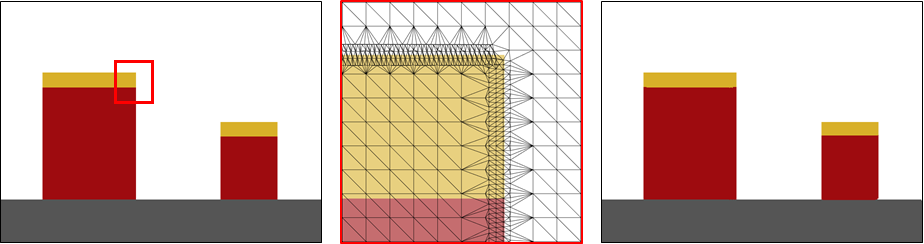}%
	\vspace{-0.2cm}
  \caption{\emph{Left:} Synthetic 2D scene, 
	colors indicate ground (\emph{gray}), building (\emph{red}) and roof (\emph{yellow}). \emph{Middle:} Control mesh. \emph{Right:} Example reconstruction. 
	}\label{fig:2d_lagrange}
\end{minipage}
%
\hspace{0.1cm}
\begin{minipage}{.385\textwidth}
  \vspace{2.5mm}
	\small
  \centering
	\begin{tabularx}{\textwidth}{X|c|c|c}
  		{\bf overall\ acc.\ [\%]} & Tetra & Octree & MB \\ \hline
 		 Scene 1 & 84.0 & 83.9 & 82.5 \\
  		Scene 2 & 92.5 & 92.8 & 89 \\
  	\end{tabularx}
		\vspace{-0.2cm}
  \captionof{table}
	{Quantitative comparison with octree model \cite{BlahaVRWPS16}
	 and MultiBoost input data.
	}\label{tab:verified}
\end{minipage}
\vspace{-6mm}
\end{figure}

\begin{figure}[t]
\begin{center}
\vspace{-0.9mm} 
   \includegraphics[width=0.885\linewidth]{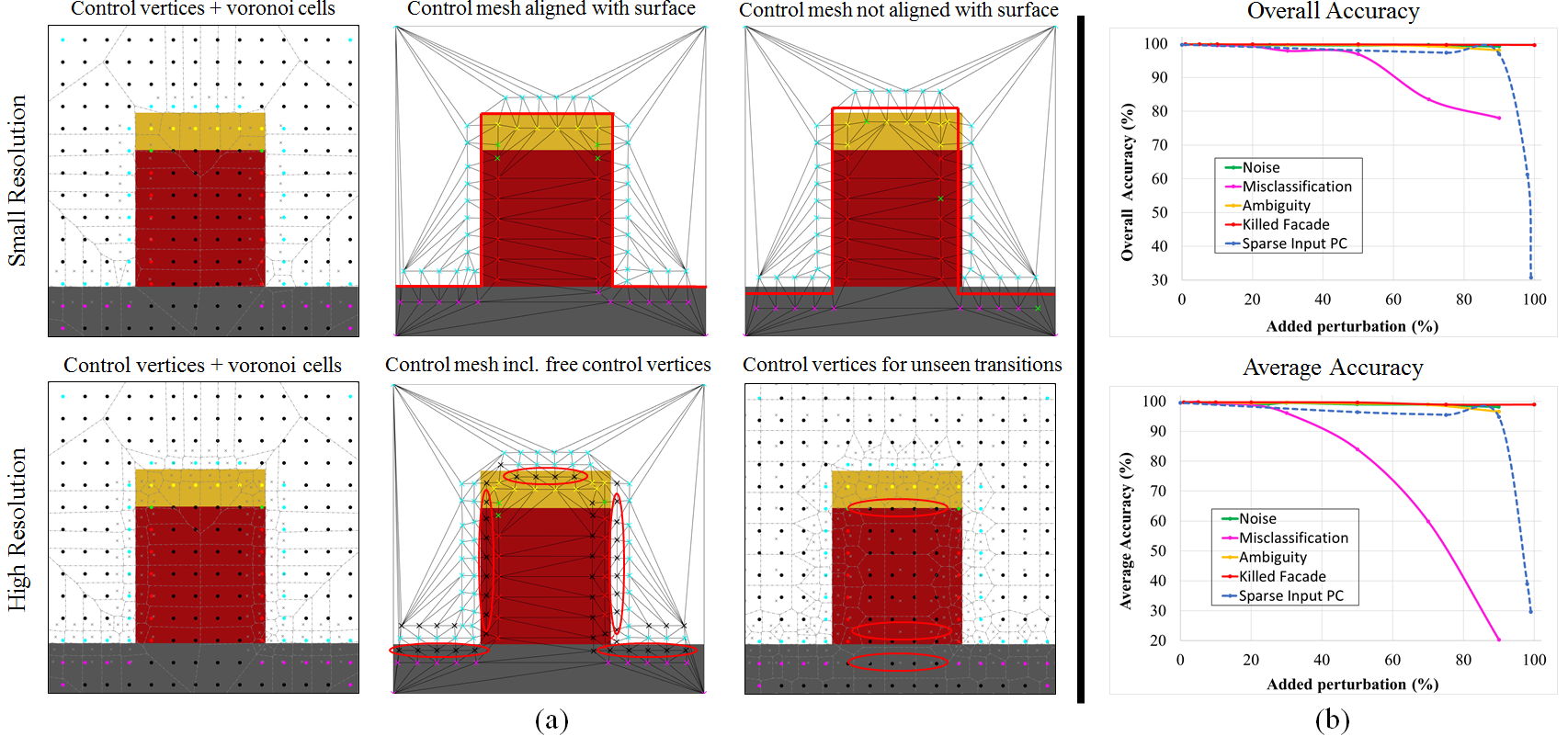}
\end{center}
   \vspace{-6.5mm}
   \caption{(a) Illustration of the control mesh foundation. Dots represent values of the datacost grid and crosses the \emph{control vertices}. Voronoi cells of the \emph{control vertices} are depicted with dashed grey lines and the \emph{control mesh} with a solid black line. Colors indicate ground (\emph{purple}), building (\emph{red}), roof (\emph{yellow}), free space (\emph{cyan}) and no datacost (\emph{black}). (b) Quantitative evaluation of Lagrange FEM method \wrt different degradations of the input data.}
\label{fig:voronoi_tri_datacost_accuracy}
\vspace{-5.15mm}
\end{figure}

\textbf{2D Lagrange results.}
Fig.~\ref{fig:2d_lagrange} illustrates the result we obtain
in a \emph{perfect} setting.
The original 2D image serves as ground truth for our quantitative evaluation.
%
In this baseline setting, our method achieves
$99.8\%$ of \emph{overall accuracy} and $99.7\%$ of \emph{average
accuracy}, confirming the soundness of our Lagrange discretisation.
%
In order to evaluate how our model behaves in a more realistic setting,
we conduct a series of experiments where we incrementally add
different types of perturbations.
Our algorithm is tested against:
%
\emph{(i)} noise in the initial 2D point cloud, respectively depth maps,
\emph{(ii)} wrong class probabilities and
\emph{(iii)} ambiguous class probabilities of random subsets,
\emph{(iv)} missing data, \eg deleting part of a facade to simulate unobserved areas,
\emph{(v)} sparsity of the initial point cloud.
Fig.~\ref{fig:voronoi_tri_datacost_accuracy}\textcolor{red}{b}
illustrates the influence of defective inputs.
Under reasonable assumptions on the magnitude of the investigated
perturbations, we do not observe a significant loss in accuracy.
%
%
The reconstruction quality starts to decrease if
more than half of the input data is misclassified
or if the input point cloud is excessively sparse, meaning that $>$50\% of
the input is wrong or nonexistent. Average accuracy is naturally more
sensitive, due to the larger relative error in small classes.
\parspc

\textbf{Influence of the control mesh.}
%
%
Recall from \eqref{eq:datacost} that the data cost of a \emph{control
vertex} $v \!\in\! V$ is approximately equal to an integral of $\rho$
over its respective Voronoi area (\cf
Fig.~\ref{fig:voronoi_tri_datacost_accuracy}\textcolor{red}{a}, left).
%
%
%
Therefore, but also because of the sign change in \eqref{eq:datacost},
vertices close to the surface receive small cost values and are
mainly steered by the regulariser, \ie these vertices realise a
smooth surface.
On the other hand, vertices that integrate only over areas with
positive or negative sign determine the inside/outside decision, but
are more or less agnostic about the exact location of the surface.
We conclude that a sufficient amount of \textit{control vertices}
should lie within the band
$[\hat{d}-3\epsilon;\hat{d}+3\epsilon]$ defined by the truncation
of the cost function around the observed depth $\hat{d}$
(\cf \eqref{eq:datacost} and \cite{HaneZCAP13}).
Ideally \textit{control vertices} are equally distributed along each line-of-sight
in front and behind the putative match
(\cf Fig.~\ref{fig:voronoi_tri_datacost_accuracy}\textcolor{red}{a}, middle column).
Undersampling within the near-band can lead to smooth, but inaccurate results
(\cf Fig.~\ref{fig:voronoi_tri_datacost_accuracy}\textcolor{red}{a}, top right).
Unobserved transitions, \eg \emph{building-ground} or \emph{roof-building},
can also lead to problems if the affected simplices are too large.
To mitigate the effect, we add a few vertices (\eg, a sparse regular grid)
on top of the \emph{control mesh} 
(\cf Fig.~\ref{fig:voronoi_tri_datacost_accuracy}\textcolor{red}{a}, bottom row).
Finally, oversampling each line-of-sight in order to increase the resolution of
the \emph{control mesh} is not recommended, the right spacing is determined by the
noise level and $\epsilon$ and $k$, chosen in \eqref{eq:datacost}.
%
%

To conclude, it is an important advantage of the FEM framework that
additional vertices can be inserted as required, without changing the energy.
In future work we will use this flexibility to develop smarter \emph{control
meshes}, possibly as a function of the \emph{local} noise level.


\begin{figure}[h]
\vspace{-2mm}
\begin{center}
   \includegraphics[width=0.96\linewidth]{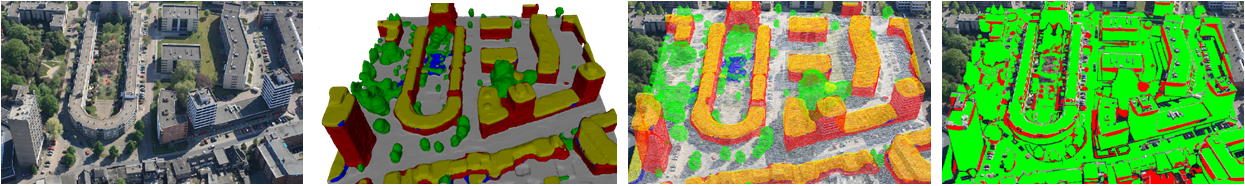}
\end{center}
   \vspace{-6mm} \caption{Quantitative evaluation of Scene 1 from
   Enschede. \emph{Left:} One of the input images. \emph{Middle left:}
   Semantic 3D model. \emph{Middle right:} Back-projected labels
   overlayed on the image. \emph{Right:} Error map, misclassified
   pixels are marked in red.}
\label{fig:3d_evaluation}
\vspace{-4mm}
\end{figure}

\parspc
\textbf{3D Lagrange results.}
To test our algorithm on real world data, we focus on a dataset
from the city of Enschede. Complementary results for other
datasets are shown in the supplementary material.
%
%
As baseline we use \cite{BlahaVRWPS16}, the current state-of-the-art
in large-scale semantic 3D reconstruction.
Due to the lack of 3D ground truth, we follow their evaluation
protocol and back-project subsets of the 3D reconstruction to
image space, where it is compared to a manual segmentation.
As can be seen in Fig.~\ref{fig:3d_evaluation} and
Tab.~\ref{tab:verified}, the two results are similar in terms of
quantitative correctness.
%
We note that measuring labeling accuracy in the 2D projection
does not consider the geometric quality of the reconstruction within
regions of a single label. 

Figs.~\ref{fig:3d_reconstruction} and~\ref{fig:3d_evaluation} show
city-modelling results obtained from (nadir and oblique) aerial
images.
Visually, our models are crisper and less ``clay-like''.
Compared to axis-aligned discretisation schemes,
\eg \cite{HaneZCAP13,BlahaVRWPS16}, our method
appears to better represent surfaces not aligned with the coordinate axis,
and exhibit reduced grid aliasing.
Both effects are consistent with the main strength of the FEM
framework, to adapt the size \emph{and} the orientation of the volume
elements to the data.
%
%
Small tetrahedra, and vertices that coincide with accurate 3D points
on surface discontinuities, favour sharp surface details and crease
edges (\eg, substructures on roofs).
Faces that follow the data points rather than arbitrary grid
directions mitigate aliasing on surfaces not aligned with the
coordinate axes (\eg, building walls).
The freedom of a local \emph{control mesh} unleashes the
power of the regulariser in regions where the evidence is
weak or ambiguous (\eg, roads, weakly textured building parts).

As already mentioned, our FEM framework can be readily combined with
on-the-fly adaptive computation, as used in the
baseline \cite{BlahaVRWPS16}.
Compared to their voxel/octree model, adaptive refinement is
straight-forward, due to the flexibility of the FEM framework, which
allows for the introduction of arbitrary new vertices.
As a preliminary proof-of-concept, we have tested the naive refinement
scheme described in Sec.~\ref{subsec:implementation}.
We execute three refinement steps, where we repeatedly reconstruct
the scene and subsequently refine simplices that contain surface
transitions, while lowering $\epsilon$ by half.
Compared to computing everything at the final resolution, this
already yields substantial savings of 89--97\% in memory and
82--93\% in computation time, without any loss in accuracy.
Targetting $\epsilon\geq\frac{1}{\sqrt{3}}$ (measured
\wrt a bounding box of 256 units), the runtimes for the tested scenes are
1h04m--1h47m and memory consumption is 573--764 MB, on a single machine.

\vspace{-0.35cm} 
\section{Conclusion}
\label{sec:conclusion}
We have proposed a novel framework for the discretisation of
multi-label problems, and have shown that, in the context of semantic
3D reconstruction, the increased flexibility of our scheme allows one
to better adapt the discretisation to the data at hand.
Our basic idea is generic and not limited to semantic 3D
reconstruction or the specific class of regularisers.
We would like to explore other applications where it may be useful
to abandon grid discretisations and move to a decomposition into simplices.
%


{\small 
\myparagraph{Acknowledgements}: Audrey Richard was supported by SNF grant $200021\_157101$. Christoph Vogel and Thomas Pock acknowledge support from the ERC starting grant 640156, 'HOMOVIS'. 
}

\clearpage
\appendix
\section*{Supplementary Material}
\graphicspath{{./justSuppArxiv/}}
\maketitle
This document provides supplementary information to support the main
paper. It is structured as follows: Sec.~\ref{sec:input_data} gives
more information about the data and pre-processing used in our
experiments, not mentioned in the paper due to lack of space.
We hope that the added details will help readers to better appreciate
the experimental results.
In Sec.~\ref{sec:visualization} we show complementary results
obtained with the proposed Lagrange FEM method on other datasets, as
well as the full large-scale reconstruction of the city of Enschede.
Sec.~\ref{sec:proofs} contains technical details and formal proofs
that had to be omitted in the paper.
Finally, Sec.~\ref{sec:raviart_thomas} discusses our formalism for
the case of the Raviart-Thomas basis (instead of Lagrange P1), leading
to piecewise constant labels. We also show results in 2D and 3D and a
comparison to those obtained with the Lagrange basis.

\section{Input Data}
\label{sec:input_data}
\begin{figure}[h]
\begin{center}
   \includegraphics[width=0.99\linewidth]{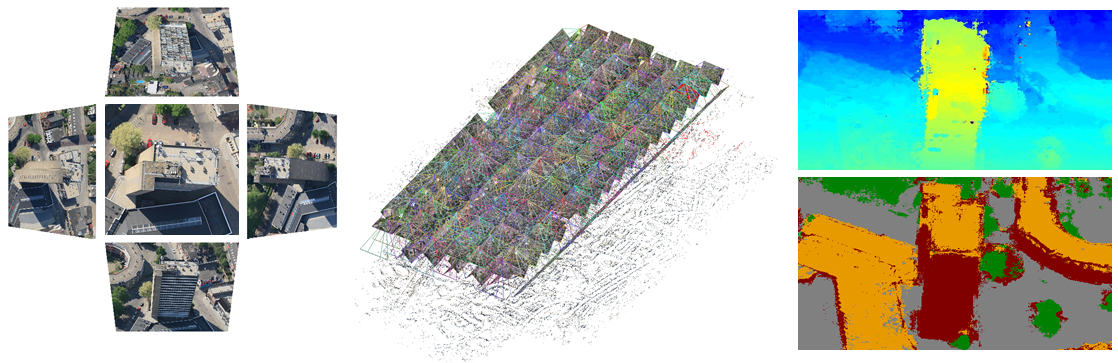}
\end{center}
   \vspace{-5mm} \caption{Input data. \emph{Left:} aerial input images
   for one position (four oblique views to the north, east, south,
   west, and a nadir image). \emph{Middle:} oriented image
   block. \emph{Right:} depth map and class probabilities (visualised
   by maximum-likelihood labels).}
\label{fig:input_data}
\end{figure}

For our real-world experiments, we start from aerial images, \cf
Fig.~\ref{fig:input_data}.
To mitigate foreshortening and occlusion, images are acquired in
a \emph{Maltese cross} configuration, with four oblique views in
addition to the classical nadir view.
We orient the images with VisualSFM \cite{wu2011visualsfm}, create
depth maps from neighbouring views with Semi-global
Matching \cite{hirschmuller2008stereo,opencv}, and predict pixel-wise
class-conditional probabilities with a MultiBoost
classifier \cite{benbouzid_jml12}. The classifier is trained on a few
hand-labeled images, using the same features as \cite{BlahaVRWPS16}:
raw RGB-intensities in a $5\times 5$ window, and $19$ geometry
features (height, normal direction, anisotropy of structure
tensor, \etc) derived from the depth map.

\section{Additional Visualizations}
\label{sec:visualization}
\begin{figure}[tb]
\begin{center}
   \includegraphics[width=0.955\linewidth]{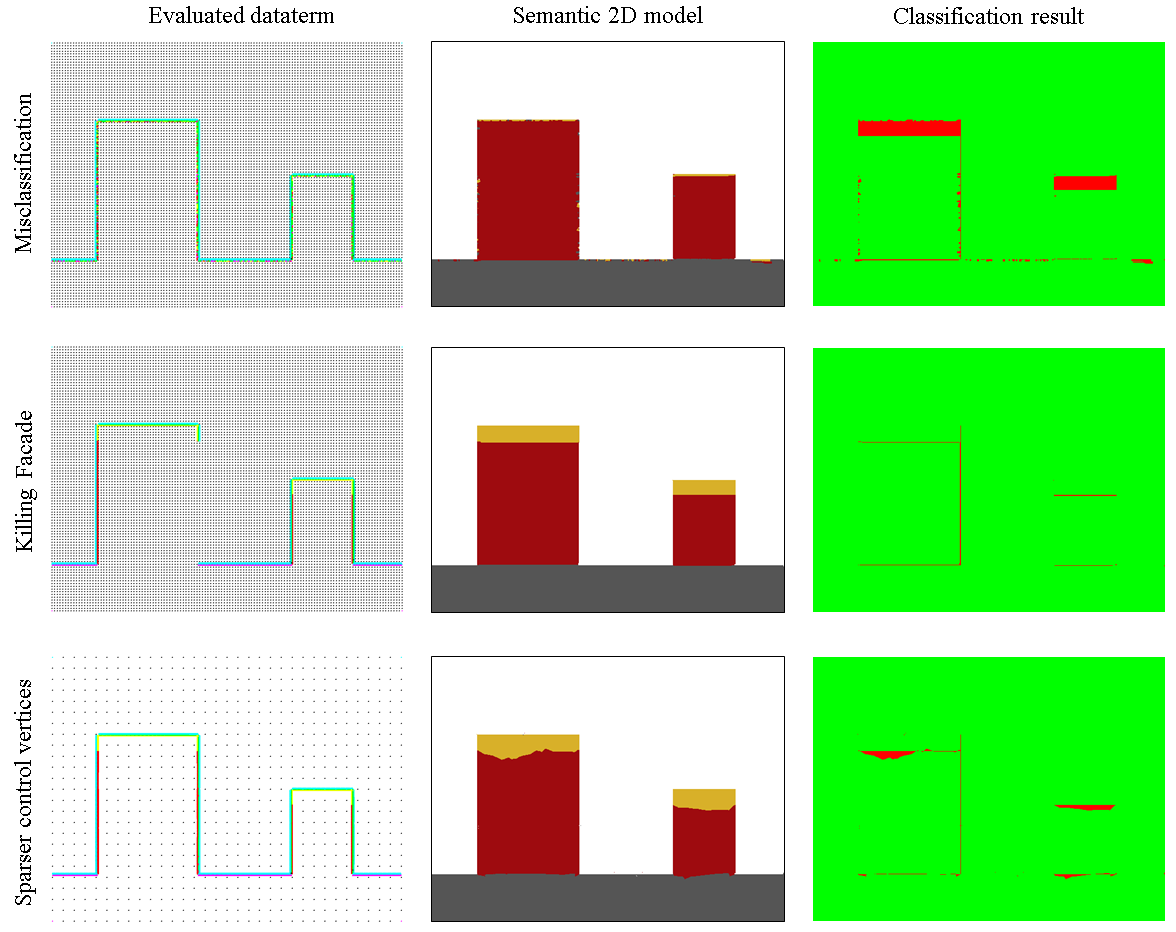}
\end{center}
   \vspace{-6mm}
	\caption{Example scenes of our 2D data set and results obtained with our Lagrange FEM method.
	\emph{Left:} Data term at vertices of the \emph{control mesh}.
	Colors for the data cost indicate: \emph{free space/empty space} (cyan), \emph{building} (red),
	\emph{ground} (pink), \emph{roof} (yellow), \emph{occupied space} (green)
	and \emph{no data cost} (black).
	\emph{Middle:} Semantic 2D model.
	\emph{Right:} Classification result, misclassified pixels are depicted in red.}
\label{fig:2d_teaser}
\end{figure}

We have tested our semantic reconstruction method on several
(synthetic) 2D and (real) 3D datasets. Here we provide additional
examples to give the reader an impression of the variety of cases
tested in our evaluation.
We apply the same prior models as for our 3D reconstructions.
We prefer flat, horizontal structures in the model for the following label
transitions: \emph{ground-freespace}, \emph{ground-building}, \emph{building-roof} and \emph{roof-freespace}.
The second prior applies to the transition \emph{building-freespace} and
prefers vertical boundaries.
Fig.~\ref{fig:2d_teaser} shows examples for different degradations of
the synthetic input (many more cases were tested).
In the top row, we simulate imperfect classifier input by adding noise
to the semantic class likelihoods.  In our experience, the method is
still able to reconstruct the geometry quite well, but sometimes
assigns the wrong label. A closer inspection reveals that, locally,
the
\emph{roof} and \emph{building} classes are confused in locations where the
class likelihoods are significantly wrong.  The global geometry and labels
in other regions remain unaffected.
The second row gives an example of missing input data, a frequent
situation in the real world, due to occlusions and constraints on
camera placement. Fortunately, missing data does not seem to greatly
challenge our method. In fact, our method is specifically designed to
work well for these cases and complete the outline, relying on the
prior assumptions about pairwise class transitions and class-specific local shape.
In the last row we utilise only a sparse control mesh, even near the
surface.  The method can still recover the geometry, but struggles to
determine the correct semantic labeling near the
(unobserved) \emph{roof-to-building} transition.
The adaptive version of our method is designed to avoid exactly that
case. It refines the \textit{control mesh} near the predicted transitions,
effectively increasing the resolution at the most promising locations.

\begin{figure}[tb]
\begin{center}
   \includegraphics[width=1\linewidth]{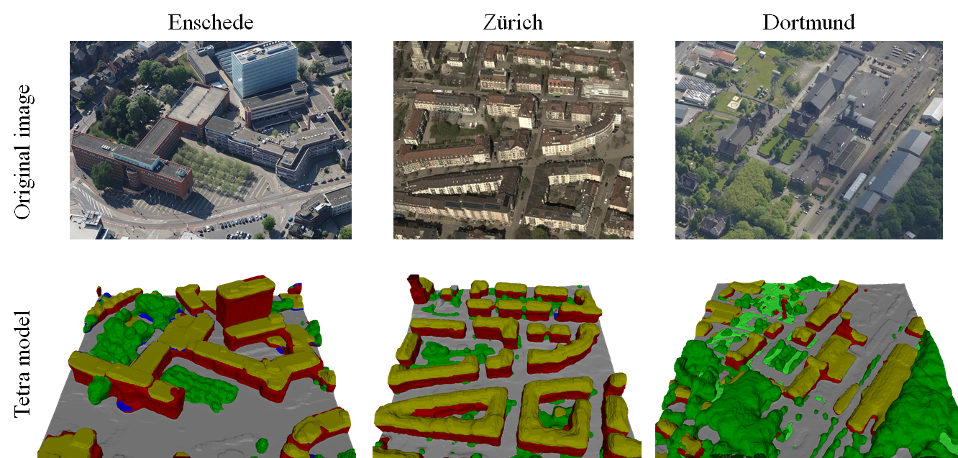}
\end{center}
   \vspace{-6mm} \caption{Additional datasets. \emph{First row:}
   Original aerial images. \emph{Second row:} Semantic 3D models
   obtained with the Lagrange FEM method for Enschede (left), Z\"urich
   (middle) and Dortmund (right). For the latter, light green denotes an additional class \emph{grass and agricultural fields}.}
\label{fig:3d_lagrange}
\end{figure}

Fig.~\ref{fig:3d_lagrange} shows city models obtained from two
additional aerial datasets (Z\"urich, Switzerland and Dortmund,
Germany), and a further patch from Enschede.
These results qualitatively illustrate that our method works for
different image sets and architectural layouts.

\begin{figure}[tb]
\begin{center}
   \includegraphics[clip=true, trim=50 0 35 0, width=1\columnwidth]{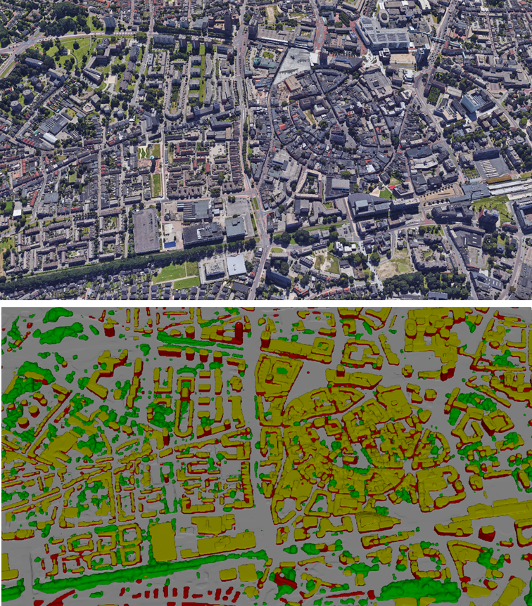}
\end{center}
   \vspace{-3mm} \caption{Large-scale semantic 3D reconstruction of
   Enschede (Netherlands), computed from aerial images with our
   Lagrange FEM method. \emph{Top:} View from \emph{Google Earth} (not
   used during reconstruction). \emph{Bottom:} Our model from matching
   viewpoint.}
\label{fig:large_scale_oblique}
\end{figure}

Finally, we show the complete semantic 3D reconstruction of Enschede.
Fig.~\ref{fig:large_scale_oblique} shows the model rendered in an
oblique view, together with the corresponding viewpoint
in \emph{Google Earth}, to illustrate its accuracy and high
level-of-detail.

\afterpage{\clearpage}

\section{Proofs}
\label{sec:proofs}
In this section we give the technical proofs promised in the main
paper, as well as further details about the optimisation.
We start with a discussion of the extension to non-metric energies,
and its consequences on the equivalence of continuous and discrete
models.
\subsection{Non-metric priors: continuous vs.\ discrete}
%

The main message of this section is that a non-metric model does not
exist in the continuous view, unless one imposes additional
constraints on the function spaces.
We briefly explain why:
Let's look at the transition boundary between two labels $i$ and $k$.
Without additional constraints, one can always introduce a zero set
with label $j$ between the two, \ie, a set with Lebesgue measure $0$
in the domain space.
If the transition costs are not metric, then the cost for the label
pair $\{i,k\}$ is potentially higher than the sum of the costs for
$\{i,j\}$ and $\{j,k\}$. Inserting the zero set will avoid that extra
cost and the energy will be under-estimated.
%
%
%
In other words, let $S$ be a segmentation of $\Omega$ into regions
$S^i$ and $S^k$, labeled with $i$ and $k$ respectively. Assume further
that their costs do not fulfill the triangle inequality \emph{\wrt} another
label $j$.
Then one can find a sequence of segmentations $S^n:=\{S_i^n, S_j^n,
S_k^n\}$ with $\displaystyle S=\lim_{n\to\infty}(S^n)$: the label $j$
disappears in the limit, such that $\displaystyle
\lim_{n\to\infty}\inf E(S^n) < E(S)$.  Hence, metric transition costs
are a necessary condition for the lower semi-continuity of the energy
functional.
%
Methods that try to resolve the issue with additional constraints on
the function space, for instance by demanding Lipshitz continuity of
the labeling functions, are an active research area, \eg
\cite{BreMas15}, but are beyond the scope of this work.
%
%
%

The above conceptual problem does have consequences for a practical
implementation:
Any discretisation of the domain will ultimately consist only of a
finite number of elements of measurable ($>0$) volume.  Thus, the
label $j$ in the example will not disappear completely from the
solution, and the computed energy matches the solution.
In practice, one can simply prescribe a minimum edge length in the
tesselation, since one cannot refine infinitely.
Note that this also constrains the Lipshitz constant of the labeling
functions; they are restricted to values between $0$ and $1$,
such that the Lipshitz constant of functions $f\in P^1(M)$ defined on the
mesh $M=\{V,F,S\}$ is bounded by
$\min_{v\in s, s\in S} ||J_v||$, \cf~\eqref{eq:gradientX}.
Because we utilise a Delaunay triangulation/tetrahedralisation of the domain
and also limit the minimal dihedral angle, a further constraint on the edge
length implies a bound on the Lipshitz constant.
%
%
Note also, our analysis implies that a discrete solution in the
non-metric setting does not have a continuous counterpart, and
consequently investigations of the limiting case, \ie, convergence
analysis after infinite refinement of the tesselation, are futile.
%
%
%



\subsection{Gradient in the Lagrange basis}\label{sec:grad_p1}
We show that gradients of functions in the P1 (Lagrange) basis are 
constant per simplex $s$ and given by:
\begin{equation}\label{eq:gradientX}
\nabla \phi_s = \sum_{v\in s} \phi_v J_{v}
\end{equation}
Here, the coefficients $\phi_v\in \mathbb{R}$ and
$J_{v}\in\mathbb{R}^d$ denote a vector of length $\frac{|f_v|}{|s|d}$,
normal to the face $f_v$ opposite to vertex $v$, and pointing inwards
towards the center of the simplex.
Recall that $|f_v|$ is the area of face $f_v$ and $|s|$ is the volume
of simplex $s$.

The gradient can be obtained with basic algebra.
First, notice that the gradient of $\phi_s$ in (\ref{eq:gradientX}) has
to fulfill $\langle v_l-v_k, \nabla \phi_s \rangle = \phi_{v_l}-\phi_{v_k}$,
meaning that integration along the edge leads to the respective change
in $\phi_s$.  After collecting a sufficient number of linear equations
of this form, one can directly solve the resulting linear system.
Since $J_{v}$ is, by definition, orthogonal to all edges that do not
involve vertex $v$, we arrive at \eqref{eq:gradientX}.

Formally, we pick one vertex $v$ of simplex $s$ and compile for
$l=1\ldots d$ ($v_l\neq v$) equations of the form $\langle v_l-v,
\nabla \phi_s \rangle = \phi_{v_l}-\phi_v$. By construction,
$\scal{J_{v_l}}{v_k-v} = \delta_{k=l}$.
The vector $J_{v_l}$ is normal to face $f_{v_l}$. The scalar product of
the edge $(v_l-v)$ and the normal is the ''height'' within the
simplex, so with the chosen scaling of $J_{v_l}$ we have
$\scal{J_{v_l}}{v_l-v_k} = 1$ for any $k\neq l$.

Thus multiplying each side of our equation system by a matrix with the
vectors $J_{v_l}, l=1\ldots d$ as columns leads to: $\nabla \phi_s
=\sum_l J_{v_l} (\phi_{v_l}-\phi_v)$. If we can show that $\sum_l
J_{v_l} = - J_v$, then we arrive at the desired expression
\eqref{eq:gradientX}. For $v_k,v_j\neq v$, $\scal{\sum_l
  J_{v_l}}{v_j-v_k} = \scal{J_{v_l}}{v_j} -\scal{J_{v_k}}{v_k} = 0$
and $\scal{\sum_l J_{v_l}}{v_k-v} = \scal{J_{v_k}}{v_k-v} =1$.  All
equations are also fulfilled by $J_v$ in place of $\sum_l J_{v_l}$,
which concludes the proof.



\subsection{Data Term for Lagrange basis}
We again start from the ideas in the main paper. We have to convert
the continuous data costs $\rho^i$ into discrete form (in a practical
implementation, ``continuous'' means that the cost can be evaluated at
any $z\in\Omega$).
In our basis representation, we can get discrete cost values for the
basis elements by convolving the continuous cost with the respective
basis function. For simplicity, we consider the P1 basis function
here.  Thus, we seek a cost per vertex $\rho^i_v$. In detail we
obtain:
\begin{align}\label{eq:dataterm_long}
\int_\Omega x^i(z) \rho^i(z) \dz =& \sum_s \int_s x_s^i(z) \rho^i(z) \dz =
\sum_s \int_s \sum_{v\in s} \phi_v p^{1}_{s,v} (z)\rho^i(z) \dz = \notag
\\
&
\sum_{v\in V} \phi_v \underbrace{\left( \sum_{s\in N(v)} \int_s p^{1}_{s,v}(z) \rho^i(z) \dz \right)}_{:=\rho_v^i} =
\scal {\rho^i_v} {x^i_v} .\raisetag{5mm}
\end{align}

To numerically compute $\rho^i_v$, we sample $\rho^i$ at a finite
number of locations $z\in \Omega$.  For each $z$ we determine into which
simplex $s$ it falls, and accumulate the contributions of
$\rho^i(z)$ over all $i=1\ldots m$, weighted by their barycentric coordinates.
%
The final step is to scale $\rho^i_v$ by $\sum_{s\in
  N(v)}\frac{|s|}{d}$ and divide by the sum of weights assigned to
vertex $v$. In other words, we compute the sample mean and scale it by
the area covered by the vertex.
In our current implementation $\rho$ is sampled at regular grid
points, without importance sampling.
This simple strategy is indeed very similar to the method employed in
\cite{HaneZCAP13,BlahaVRWPS16}.
%
%
There, the data cost is evaluated on a regular grid, by reprojecting
grid vertices into each image, computing the data term, and adding its
respective contribution to the grid location.
Such a ``per-voxel accumulation'' is equivalent to integrating the
data cost within the respective Voronoi-area of a vertex in the dual
grid: the latter is proportional to the number of regular samples that
fall into a Voronoi-cell and therefore have the respective vertex as
nearest neighbour. Hence, summing the individual contributions
directly corresponds to integrating the data term within the Voronoi
region.
%

\subsection{Grid vs.\ P1}
Here, we detail why the grid-based version with finite differences
(corresponding to \cite{HaneZCAP13}) can be seen as an approximation
of our proposed FEM discretisation with P1 basis elements, if the
vertices (cells) are aligned in a regular grid.
Without loss of generality we consider a grid of edge length $1$, and
note that in this case the gradient for a function $f:\Omega\subset
\mathbb{R}^d\rightarrow \mathbb{R}$ at a grid point x,
evaluated with forward differences becomes:
\begin{equation}\label{eq:gradientEquivalence}
\nabla f = ( f_{x+e_1}-f_x, \ldots, f_{x+e_d}-f_x )\trans 
= \sum_i e_i f_{x+e_i} - \sum_i e_i f_x
= \sum f_{x+e_i} J_{x+e_i} + f_x J_{x}, 
\end{equation}
with $e_i$ the unit vector in direction $i$.
We have used the identity
$e_i= J_{x+e_i}$, according to the definition in
\Sec\ref{sec:grad_p1}, and obtain the last equality from $\sum_l
J_{v_l} = - J_v$, \cf \Sec\ref{sec:grad_p1}.
This is exactly the formula for the gradient of the corresponding P1 function
in the simplex defined by the vertices $\{x, x+e_i\}_{i=1}^d$.
Accordingly, if implemented as finite differences, the constraints on
the dual vector field $\lambda$, see \Eq (2) from the paper,
are only checked within the respective simplex, but not in the whole domain
(1/2 of the domain in 2D; 1/6 in 3D).
Note also that, with grid-aligned vertices, the simplex in question
cannot be part of a partition of $\Omega\subset\mathbb{R}^d$, unless
$d\leq 2$: edges of adjacent faces would intersect.

\begin{figure}[tb]
\center
   \includegraphics[width=0.8\linewidth]{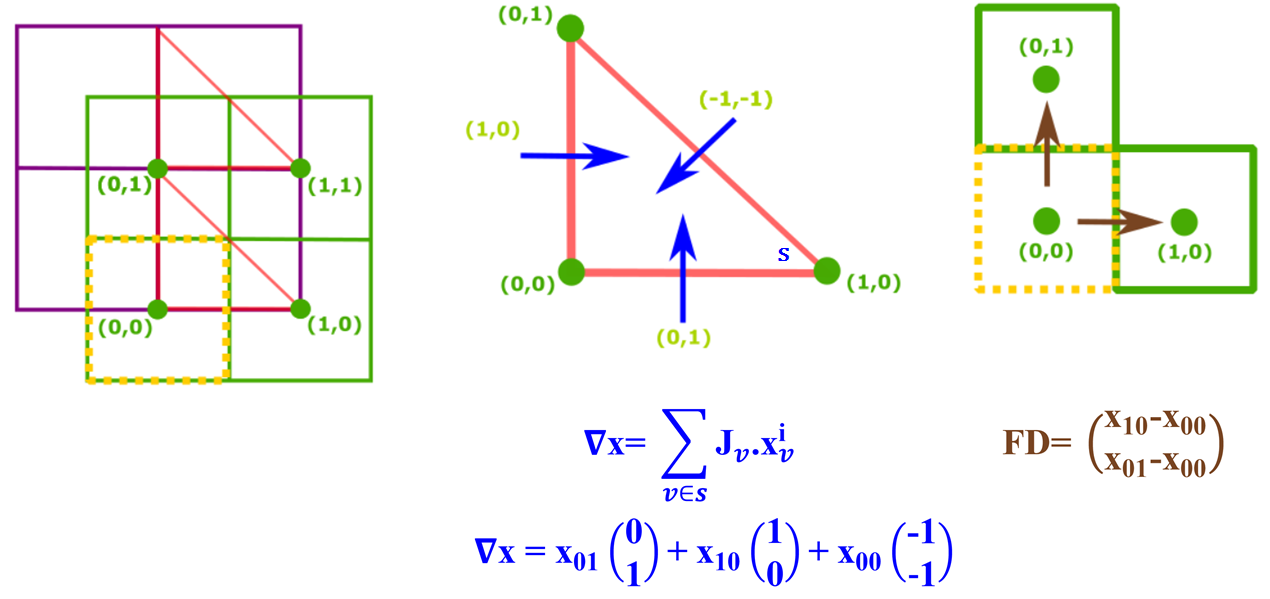}
   \caption{ \emph{Left:} Grid (\emph{green}) and simplex mesh
     (\emph{red}) cover the same domain, but are offset against each
     other.  Grid centers correspond to vertex positions.  The
     gradient in a triangle (\emph{middle}) corresponds to the
     gradient computed with forward differences (\emph{right}).  }
\label{fig:gridP1}
\vspace{-3mm}
\end{figure}

\Fig~\ref{fig:gridP1} illustrates the specific case with $d=2$. On
the left, the regular grid (\emph{green}) and the triangle
(simplex, \emph{red}).  Grid centers correspond to vertices in the
(triangle-)mesh.  The grid corresponds to the discretisation used in
\cite{HaneZCAP13}, whereas the simplex mesh is used in this paper.
The gradient of the lower left triangle for the simplex mesh
corresponds exactly to the one computed via forward differences as
shown in \eqref{eq:gradientEquivalence}.
Consequently, discretisation via finite differences is a special case
of our method, where the elements are layed out on a regular grid, and
the constraints are tested only in the upper right triangle (in the 2D
case).

\subsection{Adaptiveness}
We have stated in the paper that our formulation is adaptive, in the
sense that a hierarchical refinement of the tesselation can only
decrease the energy. We have also explained a way to find the refined
tesselation of $\Omega$, by introducing additional vertices and splitting
simplices $s$, such that no faces $f\in F$ are flipped; and we have put
forward a procedure to initialise the new variables.
Here, we formally prove that the described scheme is sound.

Let $(x^*, \lambda^*, \theta^*)$ be the solution for a triplet
$M=\{F,V,S\}$. And let $\hat{M}=\{\hat{F},\hat{V},\hat{S}\}$ be the
refined mesh with $V\subset\hat{V}$ and $\forall \hat{s}\in \hat{S},
\exists s\in S$ with $\hat{s} \cap s = \hat{s}$.
Furthermore, we define the sets $\bar{V}=\hat{V}\setminus V$ and
$\bar{S}=\hat{S}\setminus S$ to denote newly introduced vertices and
simplices.
Our construction works by induction, \ie we introduce one vertex $v$
at a time. The vertex is assumed to lie in simplex $s\in S$,
$s=\{v_k\}_{k=1}^{d+1}$, which is split into simplices
$\{\bar{s}_k\}_{k=1}^{d+1}$ with $\bar{s}_k\cap\bar{s}_l=\emptyset,
\forall k\neq l$.
By definition, vertex $\bar{v}$ has the barycentric coordinates
$\frac{|\bar{s}_k|}{|s|}$, \ie $\bar{v} = \sum_{k=1}^{d+1}
\frac{|\bar{s}_k|}{|s|} v_k$.

\begin{figure}[tb]
\begin{center}
   \includegraphics[width=0.5\linewidth]{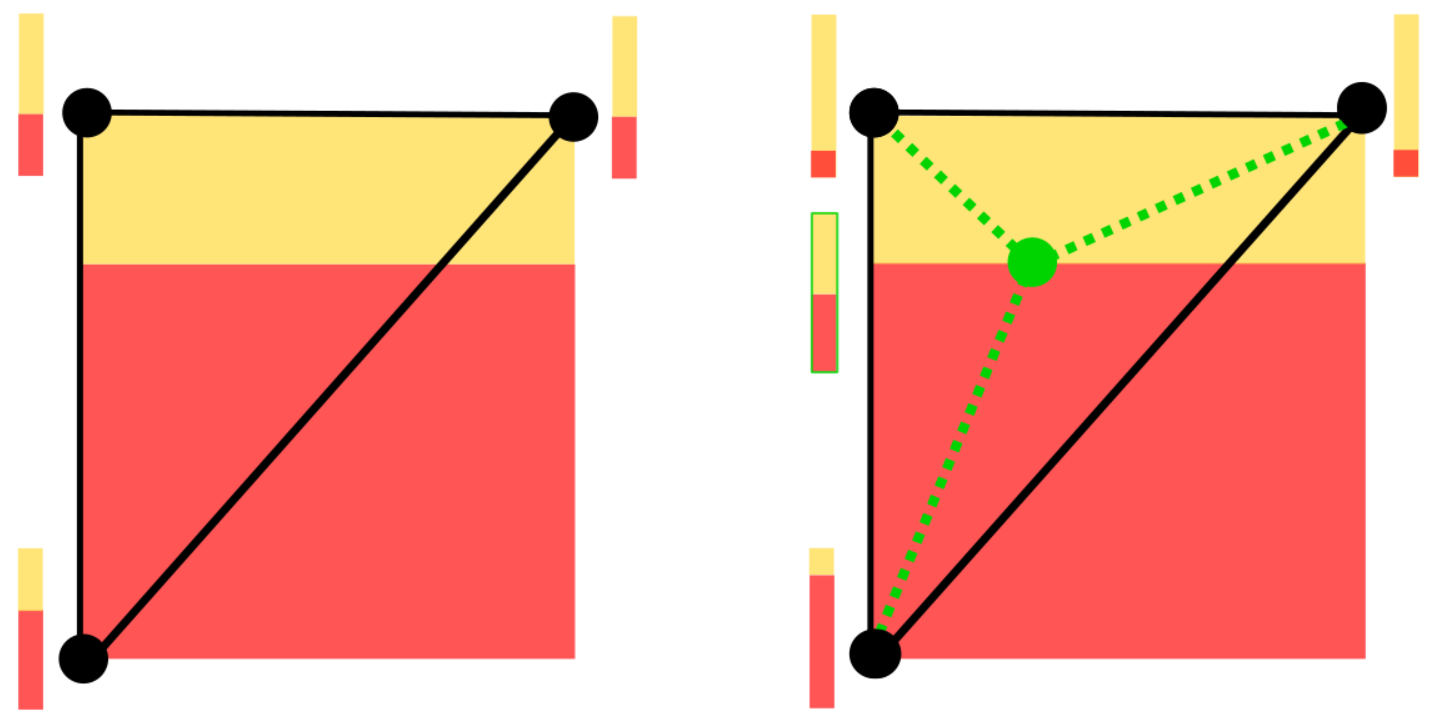}
\end{center}
   \vspace{-5mm} \caption{Updated data term after adding a new vertex.}
\label{fig:adaptiveness}
\vspace{-3mm}
\end{figure}

We initialize the labeling variables at new vertices
$\bar{v}\in\bar{V}$
via barycentric interpolation:
$x^i_{\bar{v}} = \sum_{v_k\in s} \frac{|\bar{s}_k|}{|s|} x^i_{v_k}$.
Dual variables of the new simplices $\bar{s}_k$, and also the
transition variables $x^{ij}$, are simply copied from their enclosing
simplex $s$: $\lambda^i_{\bar{s}_k} := \lambda^i_s$,
$\theta^i_{\bar{s}_k}:=\theta^i_s$ and $x^{ij}_{\bar{s}_k} :=
x^{ij}_s$.
The data terms at the new vertex $\bar{v}$, as well as at the other
vertices $v_k$ of simplex $s$, are (re)computed following
\eqref{eq:dataterm_long}, see also \Fig\ref{fig:adaptiveness}.
We call the new variables $\bar{x}, \bar{\lambda}, \bar{\theta}$ and
claim that $E_M(x^*, \lambda^*, \theta^*) = E_{\hat{M}}(\bar{x},
\bar{\lambda}, \bar{\theta})$ and that $\bar{x}, \bar{\lambda},
\bar{\theta}$ is feasible.
The latter is trivially the case (\cf \Eq (11) from the paper);
transition variables remain positive by construction, the newly
introduced labeling variables still fulfill the simplex constraints,
and they induce the same gradients in the new simplices as in the
parent simplex.
The same applies for the cost of the regulariser in the new simplices:
$ \sum_{i<j} |s| || x_s^{ij}-x_s^{ji} ||_{W^{ij}} = \sum_{i<j}
\sum_{\bar{s}_k\in s} |\bar{s}_k| ||
x_{\bar{s}_k}^{ij}-x_{\bar{s}_k}^{ji} ||_{W^{ij}}$.
More interesting is the data cost.  We introduce the notation
$\rho_{v,s}:=\int_s p^{1}_{s,v}(z) \rho^i(z) \dz$ for the data cost at
vertex $v$, originating from the integral over simplex $s$.
By induction, we need to only verify the following equality for
simplex $s$, which is split into $\{\bar{s}_k\}_{k=1}^{d+1}$:
\begin{equation}\label{eq:dataAdaptive}
\begin{split}
\sum_k x^i_{v_k} \rho_{v_k,s} = \sum_k x^i_{v_k} \sum_{j\neq k} \rho_{v_k,\bar{s}_j} + \sum_j x^i_{\bar{v}} \rho_{\bar{v},\bar{s}_j}
=
\sum_k x^i_{v_k} \sum_{j\neq k} \rho_{v_k,\bar{s}_j} + \frac{|\bar{s}_k|}{|s|} \sum_j x^i_{v_k} \rho_{v,\bar{s}_j}
\end{split}
\end{equation}
Recall we have ''old'' vertices $v_k$, $s=\{v_k\}_{k=1}^{d+1}$ and a
new vertex $\bar{v}$.
According to \eqref{eq:dataAdaptive}, we must verify:
\begin{equation}\label{eq:dataAdaptive2}
\rho_{v_k,s} \!=\! \sum_{j\neq k} \rho_{v_k,\bar{s}_j} \!+\! \frac{|\bar{s}_k|}{|s|} \sum_j \rho_{v,\bar{s}_j}
\!\Leftrightarrow\!
\int_s \!\rho(z) p^1_{s,v_k}(z) \dz \!= \!\!\int_s \!\rho(z) \sum_{j\neq k} p^1_{\bar{s_j},v_k}(z) \!+ \!\!
\frac{|\bar{s}_k|}{|s|} \sum_j p^1_{\bar{s}_j,v}(z) \dz .
\end{equation}
It is sufficient to show
\begin{equation}\label{eq:dataAdaptive3}
p^1_{s,v_k}(z) = \sum_{j\neq k} p^1_{\bar{s_j},v_k}(z) + \frac{|\bar{s}_k|}{|s|} \sum_j p^1_{\bar{s}_j,v}(z) \;\;\forall z \in s.
\end{equation}
The right hand side represents a linear function for each
$\{\bar{s}_k\}_{k=1}^{d+1}$. We can check if both sides agree on $d+1$
points in each simplex, which is easy to verify.  The locations we
check -- substitute $z$ on both sides of Eq.~\eqref{eq:dataAdaptive3}
-- are $\{v_k\}_{k=1}^{d+1}$ and $v$. These are the defining vertices
of the $d+1$ simplices $\{\bar{s}_k\}_{k=1}^{d+1}$.  Left and right
hand side vanish, except for $v$ and $v_k$.  Finally, we get
$\frac{|\bar{s}_k|}{|s|}$ for $v$ and $1$ for $v_k$ on both sides.

In our adaptive version, we directly follow the proof and split
simplices with the introduction of a single new vertex. We emphasise
again that this splitting schedule is merely a proof of concept. The
FEM discretisation allows for more sophisticated refinement
schemes, \eg, along the lines of \cite{Grinspun02}, or flipping edges
according to the energy functional, \etc.




%

\subsection{Optimisation}
%
The energy (11) from the paper is given in primal-dual form,
optimisation with existing tools is straight-forward. We apply the
minimisation scheme of \cite{chambolle11}, with
pre-conditioning \cite{Pock11}.
%
Internally, that algorithm however requires the projection onto the
Wulff-shapes $W^{ij}$, which is slightly more involved.
\subsubsection{Proxmap for the Minkowski sum of convex sets}\label{sec:proxMap}
Recall that, per label pair $\{i,j\}$, our Wulff-shapes are of the
form $W^{ij}:= \Psi^{ij} \oplus B^2_{\kappa^{ij}}$. They are the
Minkowski sum of two simple convex sets.
Recall that the $\Psi^{ij}$ encode the direction dependent likelihood of a
certain label transition.
%
In our case, all Wulff-shapes
permit a closed form projection scheme, such that we solve the
following sub-problem as proximal step, independently per simplex $s$:
%
\begin{equation}\label{eq:prox_xij}
\argmin_{x^{ij}, x^{ji}}\frac{1}{2} ||x^{ij}-\overline{x^{ij}}||^2 + \frac{1}{2} || x^{ji}-\overline{x^{ji}}||^2 +
\!\!\!\!\!\!
\sup_{w\in \Psi^{ij}\oplus B^2_{\kappa^{ij}}}\!\!\! w\trans (x^{ij}-x^{ji}) + \iota_{\geq 0}(x^{ij})+ \iota_{\geq 0}(x^{ji}) .
\end{equation}
For the following derivation we rename the two sets
$W_1:=\Psi^{ij}$ and $W_2:=B^2_{\kappa^{ij}}$.
%
%
%
In order to decouple the argument within the regulariser, we introduce
auxiliary variables $\{y_k,z_k\}_{k=0}^2$ and additional Lagrange
multipliers $\{\mu_k,\lambda_k\}_{k=0}^2$, and replace $x^{ij}$ and
$x^{ji}$ respectively:
%
\begin{equation}\label{eq:beforeOpt}
\begin{split}
\min_{x^{ij}, x^{ji},y_{k},z_{k}}\,\,\,\max_{\mu_{k},\lambda_{k}}
&
\frac{1}{2} ||x^{ij}-\overline{x^{ij}}||^2 + \frac{1}{2} || x^{ji}-\overline{x^{ji}}||^2
+
\\
\!\!\!\!\!\!\sum_{k\in\{1,2\}} \!
\sup_{w\in W_k} &
w\trans (y_k-z_k) + \iota_{\geq 0}(y_0) + \iota_{\geq 0}(z_0)
-
\sum_{k=0}^2 \lambda_k\trans  (x^{ij}-y_k) - \mu_k\trans (x^{ji}-z_k) .
\end{split}
\raisetag{0.5cm}
\end{equation}
Optimality \wrt $x^{ij}, x^{ji}$ implies:
\vspace{-0.1cm}
\begin{equation}\label{eq:optimialityCond}
x^{ij} = \overline{x^{ij}} + \sum_{k=0}^2 \lambda_k \;\;\textrm{and}\;\;
x^{ji} = \overline{x^{ji}} + \sum_{k=0}^2 \mu_k,
\end{equation}
which, after reinserting into \eqref{eq:beforeOpt}, leads to:
\vspace{-0.1cm}
\begin{equation}
\begin{split}
\min_{y_{k},z_{k}}\max_{\mu_{k},\lambda_{k}} &
\frac{-1}{2} ||\sum_{k=0}^2 \lambda_k-\overline{x^{ij}}||^2 +
\frac{-1}{2} ||\sum_{k=0}^2 \mu_k-\overline{x^{ji}}||^2
+
\\
\sup_{w_1\in W_1, w_2\in W_2}
&
w_1\trans (y_1-z_1) + w_2\trans (y_2-z_2)
+ \iota_{\geq 0}(y_0) + \iota_{\geq 0}(z_0) 
+
\sum_{k=0}^2 \lambda_k\trans  y_k + \mu_k\trans z_k \;.
\end{split}
\end{equation}
%
%
Applying Fenchel-duality yields:
\vspace{-0.1cm}
\begin{equation}
\begin{split}
\max_{\mu_{k},\lambda_{k}}\min_{z_{k}}
&
\frac{-1}{2} ||\sum_{k=0}^2 \lambda_k-\overline{x^{ij}}||^2 +
\frac{-1}{2} ||\sum_{k=0}^2 \mu_k-\overline{x^{ji}}||^2
\\&
-\iota_{W_1} (-\lambda_1) - \iota_{W_2} (-\lambda_2) -
\iota_{\leq 0}(-\lambda_0) - \iota_{\leq 0}(-\mu_0)
+ \sum_{k=1}^2 (\lambda_k+\mu_k)\trans z_k.
\end{split}
\end{equation}
The latter summand
requires $\lambda_1=-\mu_1$ and $\lambda_2=-\mu_2$:
\vspace{-0.1cm}
\begin{equation}
\min_{\mu_0,\lambda_{k}}
\frac{1}{2} ||\sum_{k=0}^2 \lambda_k\! -\! \overline{x^{ij}}||^2 \! + \!
\frac{1}{2} ||\sum_{k=1}^2 \lambda_k\! +\! \overline{x^{ji}}\! -\! \mu_0||^2
\! + \!
\iota_{W_1} (\! -\lambda_1) \! + \! \iota_{W_2} (\! -\lambda_2) \! + \!
\iota_{\geq 0}(\lambda_0) \! + \! \iota_{\geq 0}(\mu_0).
\end{equation}
In this last form, we can apply a few iterations of block coordinate
descent on the dual variables and recover the update for $x^{ij},
x^{ji}$ from \eqref{eq:optimialityCond}.
%

\section{Raviart-Thomas basis}
\label{sec:raviart_thomas}
\subsection{Methodology}
\label{sec:raviart_thomas_method}
In this section, we show how to discretise the convex relaxation,
\Eq (2) from the paper, for the case of the Raviart-Thomas basis.
For convenience, we restate the energy:
\begin{equation}\label{eq:saddleX}
\min_{x^i} \max_{\lambda^i} \sum_i\!\!\! \int_\Omega\!\! \rho^i\!(z) x^i\!(z) \!\!+\!\! \langle x^i\!(z), \div \lambda^i\!(z) \rangle \dz,
\;\;
\textrm{s.t. } \lambda^i\!(z)\!-\!\lambda^j\!(z) \in W^{ij}\!,
\sum_{i=1}^{m} \!x^i\!(z)\!\!=\!\!1,
x^i\!(z)\!\geq\!0 .
\end{equation}

The Raviart-Thomas basis is chosen as a strong contrast to the
(preferred) Lagrange basis. With Raviart-Thomas functions, we model the dual
functions $\lambda$ in \eqref{eq:saddleX},
within our trial space.
The Raviart-Thomas $\textrm{RT}^k(M)$ basis functions describe a
$div$-conforming polynomial basis of order $k\!+\!1$, \ie the divergence
of the modeled vector field is continuous across simplices.
We again discretise on a simplex mesh $M=\{F,V,S\}$
with vertices $v\in V, v\in\mathbb{R}^d$; faces $f\in F$ defined by $d$
vertices; and simplices $s\in S$ defined by $d+1$ vertices, which
partition $\Omega$: $\cup_k s_k=\Omega, s_l\cap s_k=f_{k,l}\in F$.
%
\begin{align}\label{eq:raviartT_basis}
&
RT^0(M) := \{ p:\Omega\rightarrow \mathbb{R}^d | \phi(x):=\sum_{s\in S} \phi_s(x) \textrm{ with }
\phi_s(x):=c_s x + d_s, c_s\!\in\!\mathbb{R}, d_s\!\in \!\mathbb{R}^d, \notag
\\ &
\quad\textrm{ if } x\!\in\! s \textrm{ and } 0 \textrm{ else},
\textrm{and } \phi_s(x) \textrm{ is continuous for } x\in f_v(s) \textrm{ in direction } \nu^s_{f_v}
\}.
\raisetag{1.5cm}
\end{align}
Here, we have used $\nu^s_{f_v}$ to denote the (outward-pointing) normal of
face $f_v$ of simplex $s$. By convention the face $f_v$ is located opposite
the vertex $v$.
We construct our linear basis with functions that are defined
for each face $f_v$ in a simplex $s$, 
and can be described in a \emph{local} form as:
\begin{equation*}
\phi^{0}_{s,v}(x) := (x-v) \frac{|f_v|}{|s| d} \quad \textrm{if}\; x\in s \;\textrm{and}\; 0 \;\textrm{else},
\end{equation*}
where we again let $|f_v|$ denote the area of the face and $|s|$
the volume of the simplex.
Let $\nu^s_{f_v}$ be the normal of face $f_v$ in simplex $s$,
then the basis functions fulfill:
\begin{equation}\label{eq:property_RTX}
\langle \phi^{0}_{s,u}(x), \nu^s_{f_v} \rangle := [u=v] \;\forall x\in f_v,
\end{equation}
with $[\cdot]$ denoting the Iverson bracket.

These basis functions make up the \emph{global} function space by
enforcing a consistent orientation. For each face $f$ we
can distinguish its two adjacent simplices $s^+$ and $s^-$,
by analysing the scalar product of the vector $\mathbf{1}$ and the
normal $\nu^{s\pm}_f$ of the shared face $f$
(by convention again pointing outwards of the respective simplex).
W.l.o.g., we define $\si{f}{s^\pm}:=\sign \langle \nu^{s^\pm}_f;\mathbf{1}\rangle$,
\ie $s_f^+ \nu^{s^+}_f = s_f^- \nu^{s^-}_f$.
The global basis functions per face $f_v$ are then given by:
\begin{equation}\label{eq:RT_global_basisX}
  \phi^{0}_{s,v} :=
  \begin{cases}
    \si{f_v}{s} (x-v) \frac{|f_v|}{|s| d} \quad& \textrm{if}\; x\in s\\
    0 &\textrm{else}.
    \end{cases}
\end{equation} 
In each simplex, our vector-field $\phi_s(x)\in\mathbb{R}^d$ can then be
defined in the following manner, with coefficients $\phi_{f_v}\in \mathbb{R}$:
\begin{equation*}
\phi_s(x) := \sum_{v\in s} \phi_{f_v} \phi^{0}_{s,v} (x).
\end{equation*}
By construction, \cf \eqref{eq:property_RTX},\eqref{eq:RT_global_basisX},
the vector-field is continuous along a face $x\in f$ in
direction of the face normal $\nu_f$ (of arbitrary, but fixed orientation),
\ie for neighbouring faces $s^+$ and $s^-$ we have:
%
\begin{equation}\label{eq:RT-edgeSKP}
\langle \phi_{s^\pm}(x), \nu_f \rangle = \si{f}{s^\pm} \langle x-v^\pm,\nu_f\rangle \frac{|f|}{|s^\pm|d} \phi_{f_{v^\pm}} = \phi_{f_{v^\pm}}.
\end{equation}
Here, $v^+$ is the vertex in simplex $s^+$ opposite to the shared
face, and $v^-$ is the vertex in simplex $s^-$.
Thus, our function in $\phi$ is a RT function \emph{iff} for all
neighbouring faces $s^\pm$ we have $\phi_{f_{v^+}}=\phi_{f_{v^-}}$. In
other words, basis coefficients only exist for faces of the simplices.

The variables we are interested in are the labeling functions $x^i$,
%
which are members of our test function space, composed
of piecewise constant functions per simplex:
\begin{align}
U^0(M):= \{ u:\Omega\rightarrow \mathbb{R} | &u(x):=\sum_{s\in S} u_s(x), \textrm{ with }
u_s(x)=u_s \textrm{ if } x\!\in\! s \textrm{ and } 0 \textrm{ else}\}.
\end{align}

Before we can utilise our new basis to discretise \eqref{eq:saddleX} we
need a way to enforce the constraints on our dual variables
$\lambda^i(z)\!-\!\lambda^j(z) \in W^{ij}$ for all $z\in\Omega$.
%
It is sufficient to enforce the constraints on the dual functions
$\lambda$ in \eqref{eq:saddleX} only at the
face midpoints $z_{f_v}:=1/d\sum_{w \in f_v} w$ of faces $f_v\in s$.
This ensures the constraints are also valid for any point in the
simplex $s$.
Because the Wulff shapes are convex,
it is sufficient to prove that a vector field
$\phi(x)\in {RT}^0(M), \phi(x):=\sum_{s\in S} \phi_s(x)$
at any point $x\in s$ can be written as a convex combination of the values
at the face midpoints:
\begin{equation*}
\phi_s(x) = \sum_{f_v\in s} \alpha_{z_{f_v}} \phi(z_{f_v}), \textrm{with} \sum_{f_v\in s} \alpha_{z_{f_v}}=1.
\end{equation*}
After some elementary algebra it turns out that, if $x=\alpha_i v_i$, then
$\alpha_{z_{f_v}}:=(1-d\cdot\alpha_i)$ encode this convex combination.
%
%
Furthermore, the value of $\phi_s$ at a location $x\in s$ can
be found by linear combination of basis coefficients at the vertices
of $s$:
 \begin{align}\label{eq:project_matrix}
   \phi_s(x) = \sum_{v\in s} \si{f_v}{s} (x-v) \frac{|f_{v}|}{|s|d} \phi_{v}.
 \end{align} 

\subsection{Discretisation}
\label{subsec:discretisationRT}
With these relations, we can discretise the energy \eqref{eq:saddleX}
for labeling functions $x^i\in U^0(M)$ and dual vector-field
$\lambda^i\in RT^0(M)$.
First, we convert the continuous data costs $\rho^i$ into a cost per simplex
$\rho^i_s$, which can again be achieved by convolving the cost
with the respective (per simplex constant) basis function:
$\rho^i_s := \int_s u_s(z) \rho^i(z)\dz = \int_s \rho^i(z)\dz$.
In practice, the integral is computed via sampling.
Next, we  discretise the second part of our energy
with the help of the divergence theorem and \eqref{eq:RT-edgeSKP}:
\begin{align}\label{eq:discDivX}
\int_\Omega \!\! x^i(z) \div{}\lambda^i(z) \dx \!=
\!\! \sum_{s\in S}\int_S\! x^i_s \div\lambda(z) \dx 
\!\!=\!\!
\sum_{s\in S}\int_{\partial S} \!\!\! x^i_s \langle \lambda(z), \nu(z) \rangle \dz
\!\!=\!\!\!\!\!\!
\sum_{v\in s,\,s\in S}\!\!\!\! x^i_s \lambda^i_{f_v} |f_v| \si{f_v}{s}
\end{align}
As shown, we need to verify the constraints only at face midpoints $z_{f_v}$.
The vectors $\lambda_s(z_{f_v})$ are linear in the basis coefficients for any
$z\in\Omega$, and the discretised version of \eqref{eq:saddleX} becomes
\begin{equation}\label{eq:discretisedRT_pre}
  \min_{x^i}\max_{\lambda^i} \sum_{s\in S} \rho_s^i x^i_s + \!\!\!\!\!\!\!\!
  \sum_{v\in s,s\in S}\!\!\!\! x^i_s \lambda^i_{f_v} |f_v| \si{f_v}{s}, \; \textrm{  s.t. }
  \lambda_s^i(z_{f_v})\!-\!\lambda_s^j(z_{f_v})\in W^{ij}, \;x^i_s\in\Delta \; \forall i\!<\!j,v\!\in\!s, s\!\in\!S.
\end{equation}
Here, we let $\Delta$ encode the unit simplex.
Finally, for every simplex $s$, we replace the constraint set
$\sum_{i<j}\lambda_s^i(z_{f_v})\!-\!\lambda_s^j(z_{f_v})\in W^{ij}$ in the same
manner as for the Lagrangian basis. We introduce auxiliary variables
and Lagrange multipliers $y_{s,f_v}^{ij}, \forall i<j$, and exploit
Fenchel-Duality to obtain
\begin{equation}\label{eq:stepRT}
  \begin{split}
  \max_{\lambda_s^{i}}\min_{y_s^{ij}} &
  \sum_{v\in s}\sum_{i<j}
 || y_{s,f_v}^{ij} ||_{W^{ij}} \!-\!
  \sum_{v\in s}\sum_i \langle \lambda_s^i(z_{f_v}), \sum_{j:i<j} y_{s,f_v}^{ij} - \sum_{j:j<i} y_{s,f_v}^{ji} \rangle =
\\
\max_{\lambda_s^{i}}\min_{y_s^{ij}} &
\sum_{v\in s}\sum_{i<j}
|| y_{s,f_v}^{ij} ||_{W^{ij}} \!-\!
\sum_{v\in s}\sum_i
\lambda^i_{f_v} \!\left( \!\frac{|f_v|\si{f_v}{s}}{|s|d}
\left[
    \sum_{\bar{f}\in s} (z_{\bar{f}}\!-\!v)\trans \!
    \left( \sum_{j:i<j} y_{s,\bar{f}}^{ij} \!-\! \sum_{j:j<i} y_{s,\bar{f}}^{ji} \right) \!\!
\right]\!\!
\right).
  \end{split}
\end{equation}
%
Furthermore, recall that we use Neumann conditions at the boundary of
$\Omega$,
which translates into coeffients $\lambda^i_f=0, \;\forall f\in\partial \Omega$.
%
Combining \eqref{eq:discretisedRT_pre} and \eqref{eq:stepRT}, we get
the (metric) energy for the Raviart-Thomas discretisation:
\begin{equation}\label{eq:discretisedRT}
\begin{split}
  \min_{x^i,y^{ij}}\max_{\lambda^i}
  & \sum_{s\in S} \sum_i \rho_s^i x^i_s + 
 || y_{s,f_v}^{ij} ||_{W^{ij}} + \iota_\Delta(x^i_s)
\\\raisetag{-1cm}
\!+\!&
\sum_{v\in s}\sum_i
\lambda^i_{f_v} |f_v| \si{f_v}{s} \!\left(
x^i_s -
\!\frac{1}{|s|d}
\left[
    \sum_{\bar{f}\in s} (z_{\bar{f}}\!-\!v)\trans \!
    \left( \sum_{j:i<j} y_{s,\bar{f}}^{ij} \!-\! \sum_{j:j<i} y_{s,\bar{f}}^{ji} \right) \!\!
\right]\!\!
\right)
%
%
%
\end{split}
\end{equation}
To extend it to non-metric pairwise costs, as in the Lagrangian case,
we need to impose additional assumptions.
One possibility is to utilise basis functions for the dual variables,
which are continuous in all directions at the faces.
In that case, it is only necessary to check the constraints at the faces and not
for each face in each simplex, \ie the variables for $y^{ij}_{s^+,f}$ and
$y^{ij}_{s^-,f}$ merge into one set.
Another possibility is to only force the normal component along the
faces of $\lambda$ to be contained in the Wulff-shapes.  In this
direction, RT is already continuous and the Lagrange
multipliers $y$ can be merged.  This line of attack leads to a scheme
that is remarkably similar to belief propagation on a Markov random
field, in the sense that the discretisation lacks a continuous
counterpart to begin with, and may lead to stronger grid artifacts.
We stop at this point and leave an investigation of such models
to future work.

\subsection{2D results}
\label{sec:raviart_thomas_2Dresults}

\begin{figure}[tb]
\begin{center}
   \includegraphics[width=1.0\linewidth]{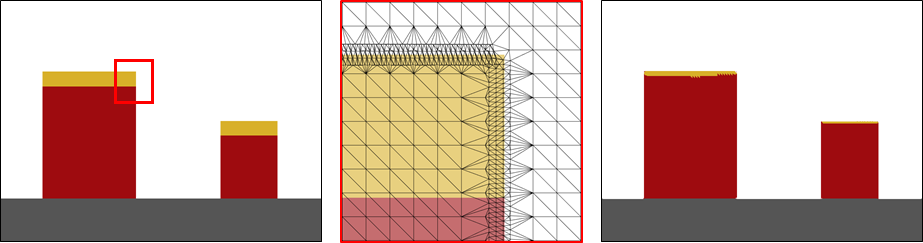}
\end{center}
   \vspace{-3mm} \caption{\emph{Left:} Synthetic 2D scene. Colors
     indicate ground (\emph{gray}), building (\emph{red}) and roof
     (\emph{yellow}). \emph{Middle:} Zoom of the \emph{control
     mesh}. \emph{Right:} Reconstructed semantic 2D model.}
\label{fig:2d_RT}
\end{figure}

Fig.~\ref{fig:2d_RT} illustrates the result we obtain with the
Raviart-Thomas FEM method (RT). We use the same (perfect) baseline
setting as for the Lagrange FEM method (P1) in the main paper. In that
setting, the RT method achieves $97.5\%$ of \emph{overall accuracy}
and $92.8\%$ of \emph{average accuracy}.
While these results confirm that also the RT method is sound, they
also show its limitations compared to the Lagrange basis.
Simplices not aligned with object boundaries, straddling multiple
labels, will necessarily introduce errors in the reconstruction.  Note
that we do not used edge information to guide the meshing; especially
since such information is not available for our target application,
semantic 3D reconstruction.
We refer to the 3D qualitative comparison (\cf
Sec.~\ref{sec:raviart_thomas_3Dresults}) for a more detailed analysis
of the differences between the two methods.\\

\noindent
\textbf{NB: }Further to this manuscript, the supplementary material
contains a short {\bf video}, which shows the diffusion of the
indicator function over $1000$ iterations for both proposed methods.\\

\begin{figure}[tb]
\begin{minipage}{.48\textwidth}
  \centering
\includegraphics[width=60mm]{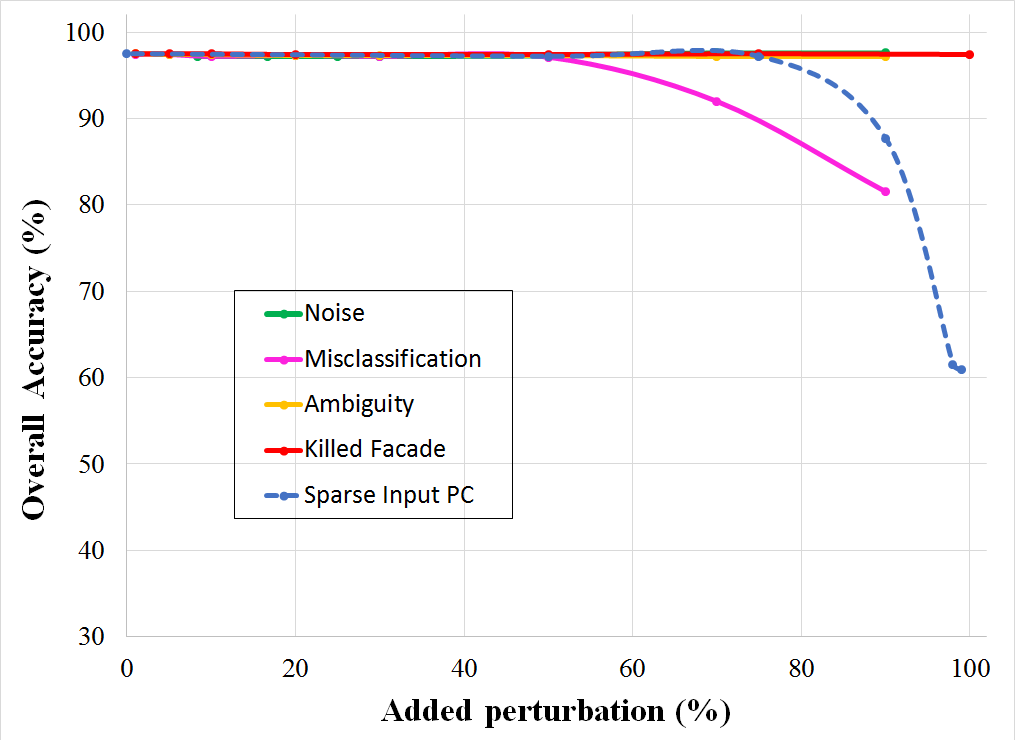}%
\end{minipage}
%
\hspace{0.1cm}
\begin{minipage}{.48\textwidth}
  \centering
\includegraphics[width=60mm]{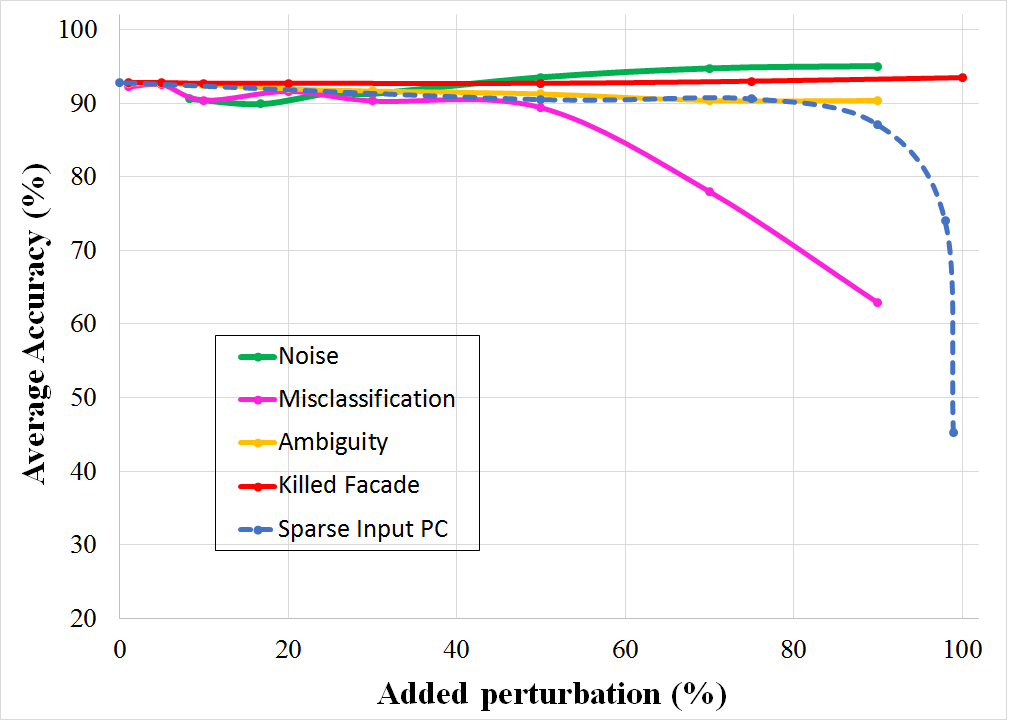}%
\end{minipage}
\vspace{3mm}
\caption{Quantitative evaluation of Raviart-Thomas FEM method \emph{w.r.t.} different degradations of the input data.}
\label{fig:ov_av_accuracy_RT}
\vspace{-3mm}
\end{figure}

We perform also the same series of experiments where we incrementally
add different types of perturbations, \cf Sec.~5 of the main
paper. Fig.~\ref{fig:ov_av_accuracy_RT} shows the corresponding
behaviour of our RT method. Generally speaking, both models shows a
similar sensitivity to defective inputs, but with a small edge for the
Lagrange method, which consistently reaches higher overall accuracy.

\subsection{3D results}
\label{sec:raviart_thomas_3Dresults}
\begin{figure}[tb]
\begin{center}
   \vspace{-5mm}
   \includegraphics[width=1\linewidth]{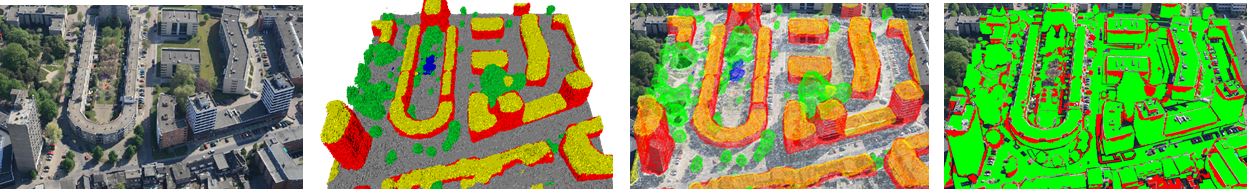}
\end{center}
   \vspace{-5mm} \caption{Quantitative evaluation of Scene 1 from Enschede. \emph{Left:} One of the input images. \emph{Middle left:} Semantic 3D model obtained with our Raviart-Thomas FEM method. \emph{Middle Right:} Back-projected labels overlayed on the image. \emph{Right:} Error map, misclassified pixels are marked in red.}
\label{fig:3d_RT_evaluation}
\end{figure}
\begin{figure}[hb]
\vspace{1em}
\center
  \begin{tabular}{cc}
   \includegraphics[width=0.475\linewidth]{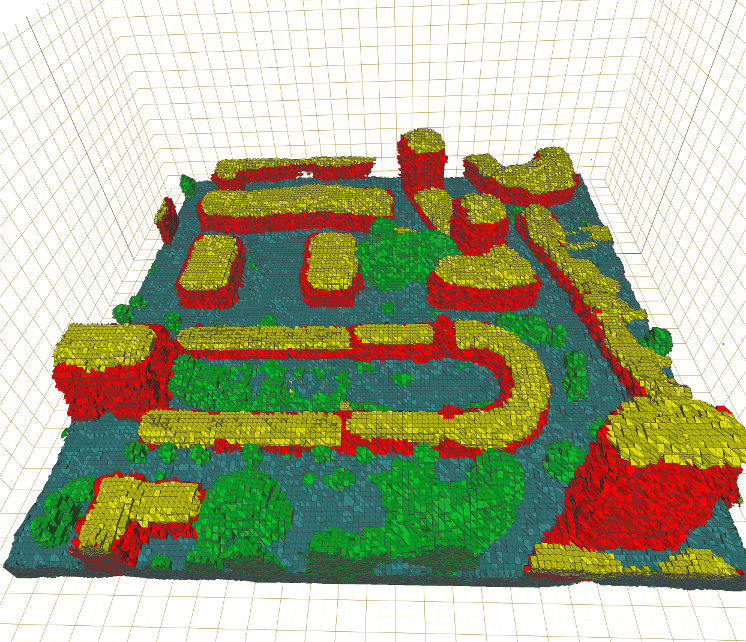} &
   \includegraphics[width=0.475\linewidth]{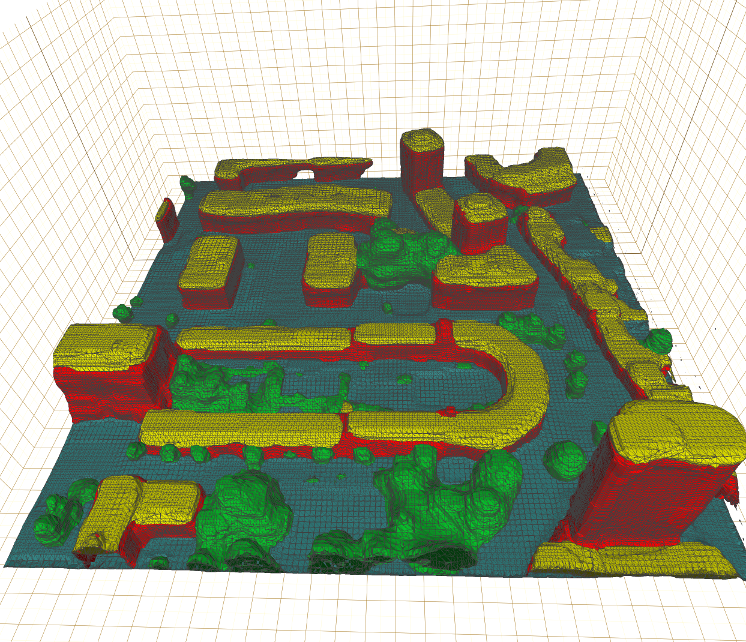}
   \\
   \includegraphics[width=0.475\linewidth]{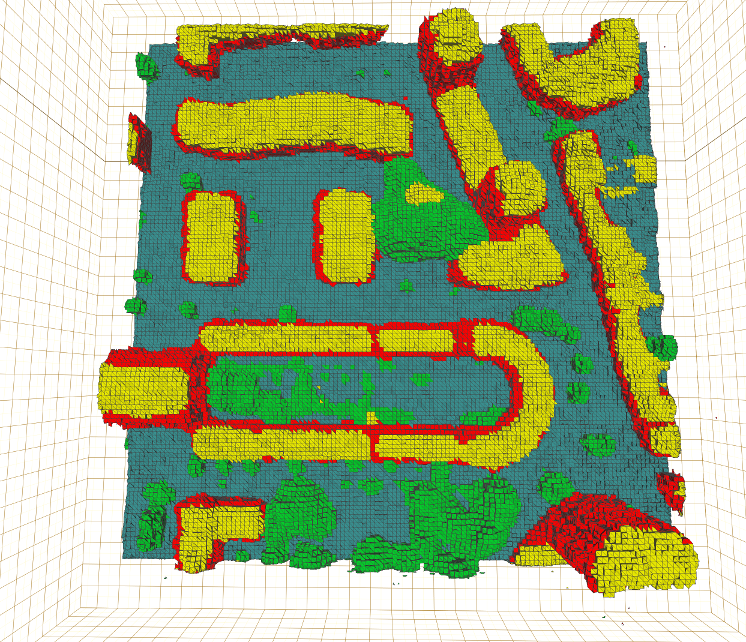} &
   \includegraphics[width=0.475\linewidth]{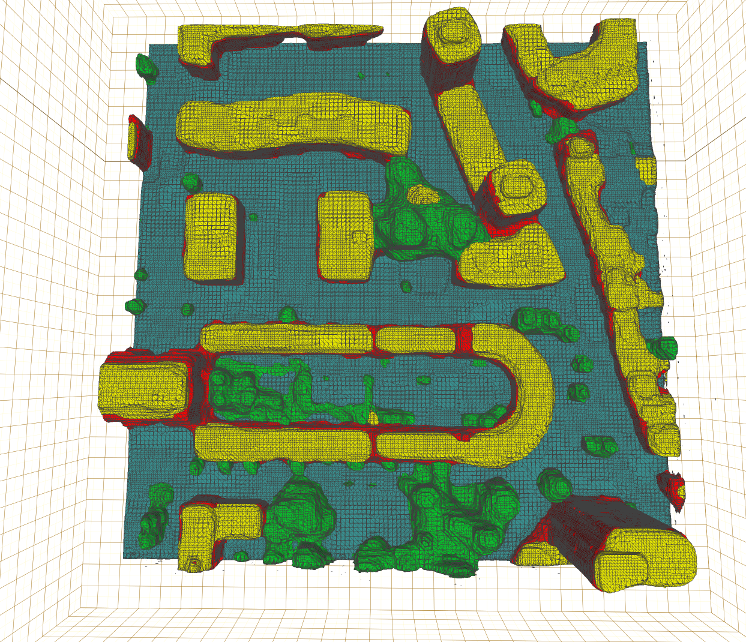}
  \end{tabular}
  \caption{Reconstruction with Raviart-Thomas (\emph{left}) and with
    the Lagrange basis (\emph{right}).  We deliberately select a low
    resolution, choose flat shading and plot mesh edges, to accentuate
    the differences. Please refer to the text for
    details. }\label{fig:qual_RT_P1}
\end{figure}

Fig.~\ref{fig:3d_RT_evaluation} shows a quantitative evaluation of the
Raviart-Thomas basis, equal to the one of the Lagrange basis presented
in the main paper. As before, the colors encode \emph{building}
(\emph{red}), \emph{ground} (\emph{gray}), \emph{vegetation}
(\emph{green}), \emph{roof} (\emph{yellow}) and \emph{clutter}
(\emph{blue}). We summarise the outcomes in
Tab.~\ref{tab:verifiedX}. The differences between the Lagrange basis
and octree are vanishingly small, on the other hand we notice a bigger
gap between the Raviart-Thomas basis and octree.
We also present a qualitative comparison of the two bases in
\Fig~\ref{fig:qual_RT_P1}.
The differences are immediately apparent, which confirms the numbers given in Tab.~\ref{tab:verifiedX}. Raviart-Thomas labels entire
simplices, so the reconstruction consists of piecewise constant
elements. On the contrary, the Lagrangian basis has the advantage that
the labeling functions are linear and can be interpreted as (signed)
distance functions, such that a smooth iso-surface can be extracted,
here done with marching tetrahedra.
Despite the piecewise constant reconstruction, the RT basis measures
metric quantities -- in contrast to, for instance, Markov random
fields, where pairwise distances between the simplices would have to
be designed explicitly to achieve similar effects.
%
\begin{table}[h]

  \centering
  \setlength\tabcolsep{5pt}
  \begin{tabular}{ l|c|c|c|c|c }

Data set & Error measure & Tetra P1 & Tetra RT & Octree & MB \\ \hline
\multirow{2}{*}{Scene 1} 
 & Overall acc. [\%] & 84.0 & 81.9 & 83.9 & 82.5 \\
 & Average acc. [\%] & 81.1 & 79.1 & 80.6 & 81.4 \\

  \end{tabular}
\vspace{2mm}
\caption{Quantitative comparison of our two proposed FEM methods with octree model \cite{BlahaVRWPS16} and MultiBoost input data \cite{benbouzid_jml12}.}
\label{tab:verifiedX}
\vspace{-0.45cm}
\end{table}

\clearpage
\bibliography{./justSuppArxiv/bib}
\end{document}


\maketitle

here we collect supplementary step by step. 


\section{Additional Details}
\label{sec:details}
\input{details_supp}


\bibliography{bib}